\documentclass{article} % For LaTeX2e
\usepackage{iclr2026_conference,times}
\iclrfinalcopy

\usepackage{latexsym}
\usepackage[T1]{fontenc}
\usepackage[utf8]{inputenc}
\usepackage{microtype}
\usepackage{inconsolata}
\usepackage{graphicx}
\usepackage{amsfonts}
\usepackage{booktabs}
\usepackage{nicefrac}
\usepackage{xcolor}
\usepackage{hyperref}
\usepackage{url}
\usepackage{enumitem}
\usepackage{amsmath}
\usepackage{comment}

\usepackage{multirow}
\usepackage{xspace}
\usepackage{caption}
\usepackage{subcaption}
\usepackage{wrapfig}
\usepackage{nicefrac}

\usepackage{float}          % [H] for figure
\usepackage{placeins}       % FloatBarrier

\usepackage[ruled,vlined]{algorithm2e}

\usepackage{amsmath,amsfonts,bm}

\def\eqref#1{equation~\ref{#1}}
\def\1{\bm{1}}

\DeclareMathAlphabet{\mathsfit}{\encodingdefault}{\sfdefault}{m}{sl}
\SetMathAlphabet{\mathsfit}{bold}{\encodingdefault}{\sfdefault}{bx}{n}

\usepackage{tabularx}

\usepackage{pifont} % for \checkmark
\usepackage{array}

\usepackage[table]{xcolor}

\newcommand\ours{\texttt{ChartAB}\xspace}

\usepackage[most]{tcolorbox}

\tcbset{
  finding style/.style={
    enhanced,
    colback=gray!10,           % light gray background
    colframe=gray!60,         % frame color
    boxrule=0.8pt,             % frame thickness
    arc=2mm,                   % rounded corners
    left=1mm, right=1mm,       % inner padding
    top=1mm, bottom=1mm,
    before skip=10pt, after skip=10pt,
    fontupper=\normalfont,     % IMPORTANT: use the normal text font
    fonttitle=\normalfont\bfseries, % bold title, but normal family/size
  }
}

\newtcolorbox{findingbox}[1][]{finding style,#1}

\title{\ours: A Benchmark for\\Chart Grounding \& Dense Alignment}
\author{Aniruddh Bansal, Davit Soselia, Dang Nguyen, Tianyi Zhou\\
University of Maryland, College Park\\
\texttt{\{ani01, dsoselia, dangmn\}@umd.edu}\\
Project: \url{https://github.com/tianyi-lab/ChartAlignBench}
}

\begin{document}

\maketitle
\vspace{-1.5em}
\textit{}\begin{abstract}
% Charts play important roles in visualization, reasoning, data analysis, and idea exchange between humans. However, existing vision-language models (VLMs) still lack an accurate perception of the details and struggle to extract fine-grained structure from charts. Such limitations in chart grounding also hinder their capability to compare multiple charts and reason with charts. 
% In this paper, we develop a novel ``\textbf{ChartA}lign \textbf{B}enchmark (\ours)'' to provide a full-spectrum evaluation of VLMs in chart grounding tasks, i.e., extracting tabular data, allocating visualization elements, and recognizing various attributes from charts of diverse types and complexities. We develop a JSON template to facilitate the calculation of evaluation metrics specifically designed for each grounding task. By applying a novel two-stage inference workflow, the benchmark can further evaluate VLMs' capability of aligning and comparing elements/attributes in two charts. 
% Our analysis of evaluations on several recent VLMs sheds novel insights on their perception biases, weaknesses, robustness, and hallucinations in chart understanding. 
% These observations expose the fine-grained discrepancies among VLMs in chart understanding tasks and indicate specific skills that need to be strengthened in existing VLMs.

Charts play an important role in visualization, reasoning, data analysis, and the exchange of ideas among humans. However, existing vision-language models (VLMs) still lack accurate perception of details and struggle to extract fine-grained structures from charts. Such limitations in chart grounding also hinder their ability to compare multiple charts and reason over them.
In this paper, we introduce a novel ``\textbf{ChartA}lign \textbf{B}enchmark (\ours)'' to provide a comprehensive evaluation of VLMs in chart grounding tasks, i.e., extracting tabular data, localizing visualization elements, and recognizing various attributes from charts of diverse types and complexities. We design a JSON template to facilitate the calculation of evaluation metrics specifically tailored for each grounding task. By incorporating a novel two-stage inference workflow, the benchmark can further evaluate VLMs’ capability to align and compare elements/attributes across two charts.
Our analysis of evaluations on several recent VLMs reveals new insights into their perception biases, weaknesses, robustness, and hallucinations in chart understanding. These findings highlight the fine-grained discrepancies among VLMs in chart understanding tasks and point to specific skills that need to be strengthened in current models.\looseness-1
\end{abstract}

% \begin{abstract}
  % The abstract paragraph should be indented \nicefrac{1}{2}~inch (3~picas) on
  % both the left- and right-hand margins. Use 10~point type, with a vertical
  % spacing (leading) of 11~points.  The word \textbf{Abstract} must be centered,
  % bold, and in point size 12. Two line spaces precede the abstract. The abstract
  % must be limited to one paragraph.
% \end{abstract}
\section{Introduction}
\label{sec:intro}

% \paragraph{Background}
Recent large multimodal models (LMMs), such as vision-language models (VLMs), have achieved remarkable breakthroughs in aligning the visual modality with language models, enabling challenging language-level reasoning on visual input signals and opening the door to a wide range of applications that naturally rely on interactions between the two modalities \citep{DBLP:conf/nips/AlayracDLMBHLMM22, li2023blip, DBLP:conf/nips/LiuLWL23a}. One critical class of applications is chart understanding and reasoning, which has broad use in finance, data science, mass media, biology, and other scientific domains where ideas and information are communicated through visualizations. In these applications, measuring numerical values in charts, comparing visual elements (e.g., bars or curves), mapping correspondences between colors, numbers, names, or markers, and recognizing attributes are essential skills for downstream tasks. Most of these tasks require accurate grounding of the structured details in charts. Moreover, dense alignment of elements across multiple charts is also a widely needed skill in practical scenarios. These challenges present new open problems for VLMs.

% \paragraph{Main Challenges}
Instead of focusing on charts, existing VLMs have primarily been pretrained and finetuned on natural images and common questions/instructions, which are not fully compatible with chart understanding tasks \citep{yao2024minicpmvgpt4vlevelmllm, laurençon2024mattersbuildingvisionlanguagemodels}. Unlike perceiving objects’ shapes, poses, and semantic meanings in natural images, accurate measurement and comparison of geometric/graphic components, understanding of their structure and layout, and manipulation of their positions and rich textual content are more critical for perception and reasoning with chart images. 
However, it remains challenging for VLMs to acquire these capabilities, often leading to hallucinations and misinterpretations in chart-centric tasks \citep{masry2022chartqa, xia2024chartx}.

Despite the recent growing interest in chart-related tasks, existing VLMs and benchmarks specifically designed for charts usually focus on simple QA tasks \citep{masry2022chartqa, masry2025chartqapro, wang2024charxiv, li2023scigraphqa}, which cannot comprehensively assess the capabilities of VLMs in grounding and understanding chart components for more general-purpose tasks. Moreover, the alignment of layouts and components across multiple charts has not been explored in previous work. Hence, there remains a lack of benchmarks dedicated to evaluating these critical skills. \looseness-1
% multi-chart grounding and dense alignment of their corresponding components. 

% \paragraph{Motivations and High-level Ideas}
In this paper, we take the first step toward systematically evaluating and analyzing general-purpose VLMs on chart grounding and multi-chart dense alignment. We formally categorize the information to be grounded in a chart into two dimensions: (1) \textbf{data}, and (2) \textbf{attributes} (e.g., colors, styles, legends, sizes, positions) that define the visualization design, components, and layout. We define the \emph{chart grounding task} as extracting both the underlying data table and the associated attributes from a chart image, and the \emph{dense alignment task} as identifying correspondences and differences between two charts. Together, these tasks represent fundamental capabilities and critical subroutines required for a wide range of chart-centric applications. 

To this end, we develop a comprehensive benchmark using pairs of similar charts to evaluate model performance on the two tasks with respect to each type of information in the two categories. To create a pair of similar charts, we perturb an existing chart by randomly modifying (1) one or a few data cells in the data table and/or (2) an attribute in the script used to generate the original chart.
To maximize the potential of VLMs and evaluate their full capabilities, we propose a multi-stage information extraction and query pipeline. In this pipeline, VLMs are first queried with a grounding task targeting specified information in each chart, followed by a comparison of the grounding results between the two charts. The pipeline leverages structured JSON templates to guide the grounding and alignment of different types of information. In addition, we introduce several novel evaluation metrics that account for the symmetry and ambiguity inherent in various types of information, thereby enabling more reliable quantitative comparisons across different VLMs.\looseness-1

Our analysis reveals the weaknesses of existing VLMs in chart perception and understanding for dense grounding and alignment. The observed errors highlight their biases and hallucinations regarding certain chart components, offering critical insights for improving VLMs. The evaluation results further show how differences across models, chart types, and queried data/attributes influence benchmarking performance. In addition, we assess the robustness of VLMs in data grounding and alignment under different attribute variations, such as changes in chart type or color schemes.

\textbf{Our contributions and novelties} are summarized as follows:
\begin{itemize}[leftmargin=1em, itemsep=0.1em]
    \item We introduce the first comprehensive benchmark, ``\ours'' to systematically evaluate VLMs' capabilities in dense grounding and alignment of data and attributes in multiple chart images. 
    \item We propose a holistic evaluation suite, including a multi-stage pipeline converting charts into JSON files with specific templates for data/attributes grounding, and a rating scheme of the grounding/alignment performance based on VLMs' answers. 
    % for questions of different components in charts.  
    \item Our evaluation and analysis of existing VLMs reveal their weaknesses in fine-grained chart understanding, highlight hallucinations, and expose biases in their vision encoders when perceiving critical chart features and structures.
    \item We evaluate VLMs' robustness on data grounding and alignment under perturbations of attributes. It provides novel insights for the design of high-quality charts. 
\end{itemize}
\section{Related Work}

\paragraph{VLMs for Charts.}
\label{related_work:vlm_for_charts}
Vision-language models have shown significant advancements in chart understanding tasks. They can be broadly classified into (1) general-purpose multimodal models and (2) chart-specialized models. General-purpose models include proprietary ones \citep{hurst2024gpt} and open-source ones \citep{abdin2024phi, chen2024expanding, liu2023improved, bai2025qwen2}. Chart-specialized models \citep{zhang2024tinychart, masry2024chartgemma, xia2024chartx, meng2024chartassisstant} demonstrate strong performance on chart benchmarks; however, they are limited by instruction tuning on specific tasks, which restricts dense-level understanding, and are further hindered by incompatible pipelines that often rely on predefined routines to handle task requirements.

% Proprietary vs open-source 

% Early methods like FigureQA \citep{kahou2017figureqa} and PlotQA \citep{methani2020plotqa} focused on traditional architecture and rule-based reasoning. Subsequent methods (DePlot \citep{liu2022deplot}, MatCha \citep{liu2022matcha}, StructChart \citep{xia2023structchart}) worked on module-based augmentation for efficient grounding of chart-data and plot-code for downstream applications. Recent methods focus on an integrated multi-task paradigm. ChartAssistant \citep{meng2024chartassisstant} utilizes mixed visual encoding and augmented pre-training for robust multi-task abilities. ChartVLM \citep{xia2024chartx} applies a difficulty-based cascading decoding mechanism to augment the model's reasoning abilities using intermediate representations. Increasingly, general-purpose VLMs have shown remarkable abilities in chart cognition and reasoning.
% \begin{related_work_connection_to_our_work}
%     The task-specific nature of chart-specialized VLMs make them infeasible for general or newer tasks. General purpose VLMs with task flexibility are hence evaluated in our benchmark experiments.  
% \end{related_work_connection_to_our_work}

% \includecomment{ChartLlaMa: instruction tuned | ChartGemma: encoder + LLM, instruction tuning | unichart: off-the-shelf visual encoder + text decoder
% TAPAS, T5: text to table, no image involved}

\paragraph{Chart Understanding Benchmarks.}
\label{related_work:chart_understanding_benchmarks}
Current chart benchmarks evaluate VLMs on specific tasks including question answering \citep{methani2020plotqa, masry2022chartqa}, summarization \citep{kantharaj2022chart}, explanation-generation \citep{kantharaj2022opencqa}. Multi-task benchmarks including ChartLlama \cite{han2023chartllama}, ChartX \cite{xia2024chartx} perform agglomeration of various modalities (like chart data, description, summary) for the downstream tasks. Recent works specifically focus on expanding QA scope to overcome increased saturation by VLMs, for example CharXiv \cite{wang2024charxiv} focuses on charts in research papers, SciGraphQA \cite{li2023scigraphqa} evaluates multi-turn QA, MultiChartQA \cite{zhu2024multichartqa} evaluates multi-hop reasoning on multiple related charts, ChartQAPro \cite{masry2025chartqapro} includes diverse visualizations such as dashboards, infographs, and flexible questions (hypothetical, unanswerable).

%  \begin{related_work_connection_to_our_work}
%     The QA driven benchmarks limit model's ability to question-specific encodings and fail to evaluate understanding of finer-level chart details.
% \end{related_work_connection_to_our_work}

% , \citep{plot2Code} explores scientific plots and more challenging use-cases

% \includecomment{ChartAssistant: existing QA + more types of charts | summarization: UniChart distillation, MMC: internet, GPT-4 text for instruction tuning data, ChartBench: Acc+ - Yes/No based and data based QA, }

% \includecomment{Failure in fine-grained analysis
%     no fine-grained analysis (QA, summarization perception)
%     evaluation - exact label match (check ChartQA, CharXiv)
%     LLM usage
% }

\paragraph{Visual Grounding.} 
The dense-level understanding abilities of VLMs have been extensively enhanced through visual grounding.
DePlot \cite{liu2022deplot} trained a transformer for image-to-CSV generation, introducing a novel table comparison method for evaluation.
StructChart \cite{xia2023structchart} proposed module-based augmentation for efficient grounding of chart data and plot code in downstream applications.
Beyond charts, the Grounded-SAM model \citep{ren2024grounded} leverages Grounding-DINO \citep{liu2024grounding} for improved dense-level open-set object tracking.
BLIP-2 \cite{li2023blip} has been widely integrated into VLMs for VQA-related tasks.
LLaVA-Grounded \cite{zhang2024llava} enables detailed text descriptions of multi-object natural images by leveraging image–text grounding for instruction tuning.

% \begin{related_work_connection_to_our_work}
%     The above works showcase inference augmentation with grounding to expand model capabilities especially for finer-level tasks requiring high degree of precision. 
% \end{related_work_connection_to_our_work}

% StructChart \citep{xia2023structchart} presented a novel data serialization for effective representation. 

% Manipulable semantic components

\paragraph{Multi-Image Reasoning.} 
Multiple benchmarks have been developed to evaluate VLMs on multi-image reasoning. MMMU \cite{yue2024mmmu} includes interleaved examples with multiple images from medical, cartoon, art, and technical domains. MUIRBench \cite{wang2024muirbench} focuses on multi-chart diagram QA but is limited to coarse-level understanding. MMIR \cite{zhao2024benchmarking} addresses chart understanding through cross-modal alignment, i.e., plotting-code correctness relative to the chart image. MileBench \cite{song2024milebench} introduces semantic understanding tasks involving text-rich images, emphasizing text extraction and comprehension in OCR, documents, and slides.

% \begin{related_work_connection_to_our_work}
%     The recent multi-image benchmarks include chart-related tasks, however they primarily focus on QA driven coarse-level reasoning thereby lacking support for fine-grained chart understanding and dense alignment.
%     % Current multi-image reasoning paradigm's chart understanding centers on traditional image-based-QA, image-to-code, image-to-OCR, interleaved text-image tasks missing evaluation of finer-level understanding of chart's plot attributes and data.
% \end{related_work_connection_to_our_work}

% \includecomment{single image: MMMU benchmark, GQA, MME, MMBench, MM-Vet, MathVista}

% MMIU \citep{meng2024mmiu} focuses on more challenging scenarios for  semantic, temporal, and spatial relationships and contains 7K images and 11K curated multiple-choice questions, revealing performance gaps in  spatial and temporal reasoning across the models tested. CV-Bench \citep{tong2024cambrian1fullyopenvisioncentric} targets spatial reasoning, object counting, and depth estimation tasks, separated into ones targeted at 2D and 3D understanding.

% CV-bench (cambrian)
% benchmarking multi-image understanding in vision-language models
\section{\ours: Chart Grounding and Alignment Benchmark}
\begin{figure}[hbt]
\centering
\vspace{-1.5em}
   \includegraphics[width=0.95\linewidth]{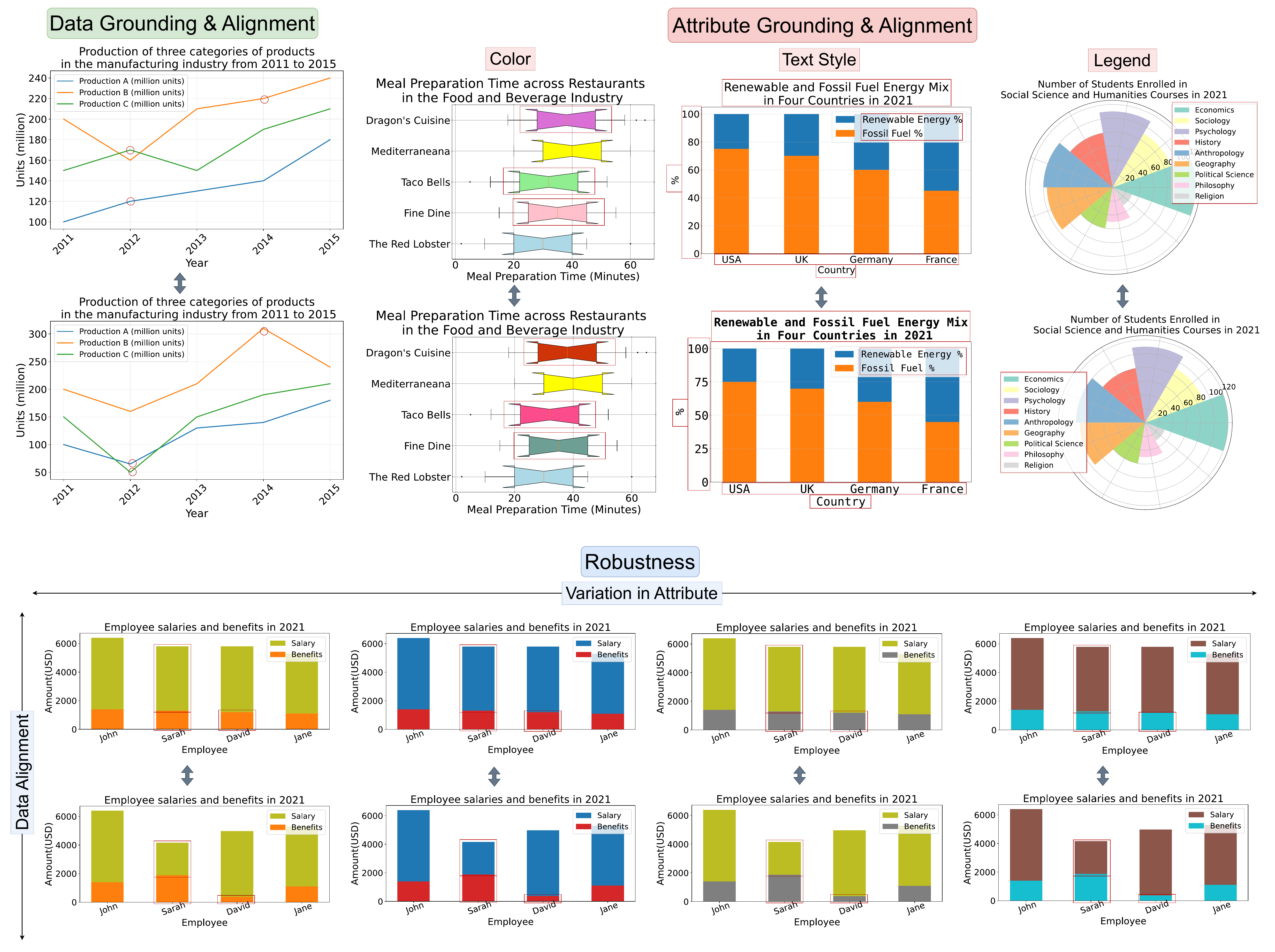}
   \vspace{-1em}
   \caption{\textbf{Examples of paired charts for \ours tasks.} \ours evaluates dense grounding and alignment capabilities of VLMs on chart images. (1) Paired charts in each \emph{Data Grounding \& Alignment} task differ in a few visualized data values. (2) Paired charts in each \emph{Attribute Grounding \& Alignment} task differ in a visualization attribute, e.g., color, legend position, or text style. (3) Each \emph{Robustness} task contains multiple variants of the same chart-pair for \emph{Data Alignment}, with different attributes (e.g., colors) across the variants.}
    \label{fig:dataset_examples}
\end{figure}

We introduce \ours, the first benchmark designed to evaluate vision-language models (VLMs) on dense level chart understanding. The benchmark focuses on three core capabilities essential to chart reasoning: (1) \textit{grounding}: extracting structured information from a single chart image, (2) \textit{alignment}: identifying fine-grained differences between a pair of similar charts, and (3) \textit{robustness}: assessing the stability of alignment performance under variations in chart appearance. 
% The evaluation of these capabilities is performed on two types of chart content: (1) \textit{data}: underlying numerical values visualized, and (2) \textit{attributes}: visual elements (e.g. color) that define chart's design. 
These capabilities serve as cornerstones for a wide range of downstream applications. We develop a novel two-stage pipeline that can isolate and rigorously evaluate them. Thereby, \ours offers a deeper diagnostic suite of VLMs' perceptual accuracy, reasoning limits, and alignment behavior in structured visual domains. \looseness-1

% We construct the \ours dataset for evaluating these capabilities. We include several examples from our dataset in Figure \ref{fig:dataset_examples}. The dataset covers diverse chart types from various domains with fine-grained ground-truth labels. 
\includecomment{downstream QA: We also evaluate the VLMs on downstream QA to analyze the impact of grounding/alignment of higher-level reasoning ability.}

% We present ChartAlignBench: the first dataset for evaluating dense-level alignment in charts across following 3 tasks: \textit{Data Alignment}, \textit{Plot-Attribute Alignment}, \textit{Robustness}. The 3 alignment tasks consist of $\sim$ 3,600, $\sim$ 2,000, $\sim$ 3,300 instances respectively. 
% For Data Alignment \& Plot-Attribute Alignment, each instance consists of pair of chart-images diverging in finer-level chart-data \& plot-attribute respectively. For Robustness, each instance contains 5 pairs of chart-images, each pair with identical chart data divergence but variation in a plot attribute across the 5 pairs. 

% \begin{figure*}[htp]
% \centering
% \vspace{-1em}
%    \includegraphics[width=0.95\linewidth]{imgs/robustness_examples.pdf}
%    \vspace{-1em}
%    \caption{Robustness examples from the \ours dataset}
%     \label{fig:dataset_robustness_examples}
% \end{figure*}

\subsection{Dataset Taxonomy and Construction}

% \subsection{Taxonomy}
% \label{subsec:taxonomy}

\begin{wraptable}[9]{r}{0.5\linewidth}
\tiny
\vspace{-2em}
% \begin{table}[htbp]
\centering
\setlength{\tabcolsep}{2.0pt} % tighter column spacing
\renewcommand{\arraystretch}{1.2} % compact row height
\caption{\textbf{Task Taxonomy in \ours}, which is composed of three types of tasks defined on different data cells and attributes. 
% evaluation category (Grounding, Alignment, Robustness) is applied over data and attributes. \looseness -1}
}
\vspace{-1em}
\resizebox{1.0\linewidth}{!}{
% \scalebox{0.9}{
\begin{tabular}{|l|c|c|c|c|c|c|c|c|}
\hline
\multirow{3}{*}{\textbf{Task Type}} 
& \multicolumn{3}{c|}{\textbf{Data}} 
& \multicolumn{5}{c|}{\textbf{Attributes}} \\
\cline{2-4} \cline{5-9}
% \cmidrule(lr){2-4} \cmidrule(lr){5-9}
& 1-Cell & 2-Cell & 3-Cell 
& Color & Legend 
& \multicolumn{3}{c|}{Text Style} \\
\cline{7-9}
& & & & & & Size & Weight & Font Family \\
% & \multicolumn{5}{c|}{} & Size & Weight & Font \\
\hline

% ---------- Grounding ----------
\textbf{Grounding} 
& \multicolumn{3}{c|}{\textbullet} % merged 3-cell columns for Data Grounding
& \textbullet & \textbullet 
& \textbullet & \textbullet & \textbullet \\ % merged text style subtasks
\cline{1-9}

% ---------- Alignment ----------
\textbf{Alignment} 
& \textbullet & \textbullet & \textbullet
& \textbullet & \textbullet 
& \multicolumn{3}{c|}{\textbullet} \\
\cline{1-9}

% ---------- Robustness ----------
\textbf{Robustness} 
& \textbullet & \textbullet & \textbullet & \multicolumn{5}{c|}{\cellcolor{gray!30}}\\

\hline
\end{tabular}}
\vspace{-1em}
\label{tab:task_taxonomy}
\end{wraptable}
% Motivated by existing criteria for dense level understanding of charts,
% \paragraph{Chart Content: Data \& Attributes} 
% \label{par:chart_content}
We construct \ours from ChartX~\cite{xia2024chartx} as the source dataset. It encompasses diverse chart types from various domains, including commerce, industry, lifestyle, society, and culture, and provides both CSV data and plotting code for each chart. 
% The dataset covers diverse chart types from various domains with fine-grained ground-truth labels
We list the taxonomy of \ours in Table~\ref{tab:task_taxonomy}. 
For each chart, we extract dense annotations of two types of fine-grained information: 
(1) \textit{Data}: The underlying data table that the chart visualizes. 
(2) \textit{Attributes}: The visual attributes that defines the appearance of the chart, e.g., \textit{color}, \textit{legend}, and \textit{text Style}. 
In particular, \textit{color} refers to the colors of the visual elements as bars, lines, or boxes in charts. 
\textit{Legend} refers to the position of the chart legend. 
\textit{Text Style} captures the textual characteristics in four chart regions: title, legend, axis labels, and axis ticks. These characteristics include textual size, weight (lightness/boldness), and font family (e.g., Times New Roman). \looseness-1

\includecomment{1. Chart pair construction (for performing the tasks)
2. Example figure, and dataset stats}

\begin{wrapfigure}[21]{r}{0.5\linewidth}
% \captionsetup{justification=raggedleft, singlelinecheck=false} % left-align caption
%  \raggedleft
\vspace{-2em}
\centering
\includegraphics[width=0.9\linewidth]{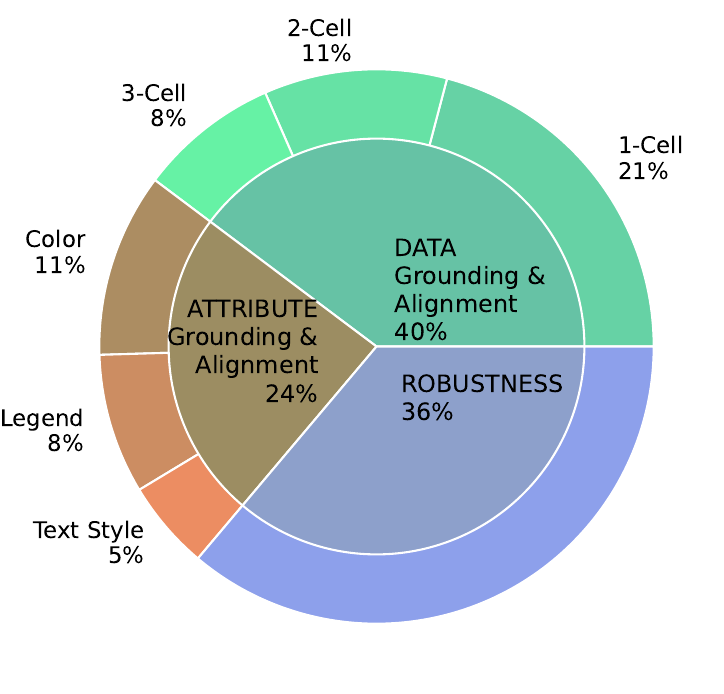}
   \vspace{-2em}
%    \caption{\textbf{Dataset statistics for \ours.} 
% The benchmark consists of 9k instances i.e. pair of chart images. 
% Pairs in \textit{Data Grounding \& Alignment} differ in one, two, or three data cells. 
% Pairs in \textit{Attribute Grounding \& Alignment} differ in appearance namely color, legend position, or text style. 
% Pairs in \textit{Robustness} task share identical data differences but introduce variations in attributes.
\caption{\textbf{Statistics of \ours.} 
\ours includes 9,000+ instances curated for tasks below: 
(1) Paired charts for \emph{Data Grounding \& Alignment} differ in one to three data cells; 
(2) Paired charts for \emph{Attribute Grounding \& Alignment} differ in color, legend position, or text style;  
(3) \emph{Robustness} task includes multiple pairs that share identical differences in data but differ in certain attributes.\looseness-1}
    \label{fig:dataset_statistics}
    % \vspace{-1em}
\end{wrapfigure}
Section~\ref{par:tasks} introduces three types of tasks built upon the dense annotations. Grounding tasks aim to extract these dense labels, while robustness tasks  evaluate grounding performance under perturbations of attributes. 
Alignment tasks introduced aim to identify the differences between two similar charts. To create pairs of similar charts, we draw an image from the ChartX, apply controlled modifications in the plotting code, and execute the code to render an variant of the original chart. Each chart’s source data (CSV file) and plotting script are provided in ChartX, ensuring precise ground-truths. 

Figure~\ref{fig:dataset_examples} provides several examples of different tasks, while Figure~\ref{fig:dataset_statistics} reports the statistics of these tasks. 
\ours covers nine diverse chart types with different data and attribute perturbations: (1) \textit{simple charts}: bar chart, bar-numbered chart, line chart, and line-numbered chart; (2) \textit{complex charts}: 3D chart, box chart, radar chart, rose chart, and multi-axes chart. More details about chart data curation are provided in \ref{app:dataset_curation}.

% We generate variations in data and attributes for charts by applying controlled perturbations to the plotting code followed by code execution to render chart images. 
% \dang{I checked the Appendix A.2 but still find it unclear how the perturbations were done. Could you give an example/figure to explain/illustrate this? What kinds of chart perturbations were considered, e.g., translating? scaling? modifying numerical values? etc. If applicable, can you provide a pseudocode to perform controlled perturbation?} 

% \subsubsection{Chart Content: Data \& Attributes}

\subsection{Evaluation Tasks}
\label{par:tasks}
% \subsection{Grounding of Single Chart}
% \label{par:chart_grounding}
% \paragraph{\textbf{Grounding of Single Chart}}

% \begin{wrapfigure}[8]{r}{0.5\linewidth}
% \centering
% \includegraphics[width = 0.98\linewidth]{imgs/task_taxonomy.pdf}
% \vspace{-1em}
% \caption{Task Taxonomy}
% \label{fig:task_taxonomy}
% \end{wrapfigure}

% sizes -> \tiny, \scriptsize, \footnotesize, \small, \normalsize,

% \end{table}

% \begin{wrapfigure}[19]{r}{0.5\linewidth}
% % \captionsetup{justification=raggedleft, singlelinecheck=false} % left-align caption
% %  \raggedleft
% \vspace{-3em}
% \centering
% \includegraphics[width=0.9\linewidth]{imgs/task_taxonomy.pdf}
%    \vspace{-2em}
% %    \caption{\textbf{Dataset statistics for \ours.} 
% % The benchmark consists of 9k instances i.e. pair of chart images. 
% % Pairs in \textit{Data Grounding \& Alignment} differ in one, two, or three data cells. 
% % Pairs in \textit{Attribute Grounding \& Alignment} differ in appearance namely color, legend position, or text style. 
% % Pairs in \textit{Robustness} task share identical data differences but introduce variations in attributes.
% \caption{}

%     \label{fig:}
%     % \vspace{-1em}
% \end{wrapfigure}

\textbf{Grounding of Single Charts }
Dense grounding of chart elements requires the extraction of precise semantic information from chart images.
% , including visualized data and visualization attributes. 
However, general-purpose VLMs are trained to mainly focus on global visual features or major objects in scenes. When applied to charts, they often fall short of perceiving the details \citep{xu2023chartbench}, which are crucial for chart reasoning. Prior works primarily evaluate VLMs' chart understanding capabilities via QA tasks, which do not fully capture their semantic grounding or reflect their cross-modal inconsistencies \citep{huang2024visual}. 
To ensure interpretable and compositional reasoning, we need to examine whether VLMs can ground the chart information in textual form. 

We formalize \textit{Grounding} as the conversion of a chart image into a structured textual representation of data or attributes.
% a chart image as input, resulting in a structured textual representation of its contents. 
As shown in Table \ref{tab:task_taxonomy}, we assess this capability through the following tasks: (1) Data Grounding, (2) Color Grounding, (3) Legend Grounding, (4) Text Style Grounding (subtasks: Size, Weight, Font Family).
\textit{Data Grounding} requires the VLM to generate a standard CSV representation of the data table.
% , including the headers (i.e., rows and columns) and cell values. 
We provide a JSON template for tasks requiring Attribute Grounding (\textit{Color/Legend/Text Style}) and prompt the model to generate a JSON representation.

Grounding the chart image into textual form isolates the model's perceptual ability from downstream prompt variation or instruction complexity. This helps build a foundation for the subsequent dense alignment and QA tasks, while also enabling failure analysis of VLM in perceiving chart components.

% As shown in table \ref{tab:task_taxonomy}, grounding is performed on the two types of chart content: (1) grounding of \textit{data} 
% It can be performed on two types of content: (1) \textit{data}: underlying data table that the chart visualizes. We prompt the model to generate a standard CSV representation of the data table, including the headers (i.e., rows and columns) and cell values; (2)  We provide a JSON template for each attribute and prompt the model to generate a JSON representation. Grounding the chart image into textual form isolates the model's perceptual ability from downstream prompt variation or instruction complexity. This helps in building a foundation for the subsequent dense alignment and QA tasks, while also enabling failure analysis of VLM in perceiving chart components.

\includecomment{ And helps avoid issues from VLM's cross-modal inconsistencies. attention bias towards text region: impacts grounding NOT QA; uniform representations - not sensitive to prompt, not an issue to sidestep using grounding, instead it encourages QnA approach}

\textbf{Dense Alignment between Two Charts }
\label{par:dense_alignment}
% \paragraph{\textbf{Dense Alignment between Two Charts.}} 
While single chart grounding evaluates a model’s perception of details in a given chart, multi-chart reasoning in practice often requires comparing similar charts to detect and analyze the differences among them. To evaluate this capability, we define a dense alignment task where the model identifies fine-grained discrepancies between two charts. Crucially, this task builds on grounded representations, allowing us to isolate and evaluate comparative reasoning for given chart pairs. As shown in our ablation studies (\ref{app:ablation_study}), direct alignment without grounding yields significantly weaker performance, highlighting the necessity of grounding for subsequent dense alignment.

% \label{par:dense_alignment_formulation}
We formalize \textit{Dense Alignment} as a comparison of two chart images that differ in local details of data or attributes. As shown in Table \ref{tab:task_taxonomy}, we assess this capability via the following tasks: (1)\textit{ Data Alignment}, (2) \textit{Color Alignment}, (3) \textit{Legend Alignment}, (4) \textit{Text Style Alignment}. \textit{Data Alignment} task is further divided into subtasks: \textit{1-cell},\textit{ 2-cell}, and \textit{3-cell}, which perform dense alignment of data for chart images that differ in 1, 2, and 3 data points, respectively. 
Each alignment task challenges the model to identify the set of divergent content and produce a structured JSON listing these differences. 

\textbf{Robustness of Data Alignment to Attribute Variation }
% \label{par:robustness}
% \paragraph{\textbf{Robustness of Data Alignment over Attribute variation}}
Using VLMs for real-world understanding of charts requires analyzing charts in diverse visual forms, i.e., diverse attributes (color/text style/legends) presence for similar types of data, often due to differing plotting tools. Moreover, past work shows the sensitivity of VLM's chart understanding under attribute changes \citep{guo2024understanding}. Hence, it motivates the evaluation of VLM's chart understanding consistency across noise, style shifts, and design variations due to variations in attributes.  

We thus formalize \textit{Robustness} of Data Alignment 
% (discussed in \ref{par:dense_alignment_formulation})
to variation in Attributes (Color/Legend/Text Style). 
To perform the task, each instance contains five pairs of chart variants created from the same pair of charts. Each pair visualizes the same source data and maintains identical data differences as the other four pairs, but their attributes (e.g., color of bars) vary across the five pairs. \looseness-1

\textbf{Effects of Dense Grounding \& Alignment on Downstream QA Tasks }
% The practical application of VLM on charts is towards diverse downstream tasks requiring complex reasoning. The grounding \& alignment ability serve as building blocks for the downstream reasoning hence form cornerstone for VLM's effectiveness on the subsequent corresponding downstream tasks. And errors in grounding/alignment tasks are common reasons behind the failures on these high-level reasoning tasks. For analyzing downstream abilities, we evaluate VLM for Question-Answering (QA) on the chart image. The QA evaluation as most widely applied downstream task (discussed in \ref{related_work:chart_understanding_benchmarks}) along with precise objective scoring motivate its selection. ChartX dataset's \cite{xia2024chartx} QA set is utilized. The QA set's questions are generated focusing on chart data with 1-word answer format with binary result which can be answered using knowledge of csv data table. 
Practical applications of VLMs on chart-related tasks often require complex reasoning, in which dense grounding \& alignment usually serve as foundational building blocks and the cornerstone of various downstream tasks. 
% for the downstream reasoning hence form cornerstone for VLM's effectiveness on the subsequent corresponding downstream tasks. 
On the other hand, grounding/alignment errors are common reasons for many reasoning failures of VLMs on charts. 
To demonstrate the importance of dense grounding/alignment skills, we evaluate VLMs on QA tasks, the most widely applied category of downstream tasks, and investigate the correlation between QA performance and the grounding/alignment quality scores. 
To this end, our study is conducted on QA tasks from ChartX~\citep{xia2024chartx} that have single-word answers derived from the grounded CSV tables.  

\subsection{A Two-Stage Evaluation Pipeline}
\label{par:multi-stage_reasoning_in_o3_for_dense_alignment_in_charts}

% The two-stage approach fundamentally decomposes the dense-alignment task into sub-tasks grounding to perform finer-level analysis. The task decomposition converts complex, finer-level reasoning into smaller steps for efficient element-wise comparisons and handling model biases. 

\begin{wrapfigure}[23]{r}{0.7\linewidth}
\centering
\vspace{-2em}
   \includegraphics[width=1\linewidth]{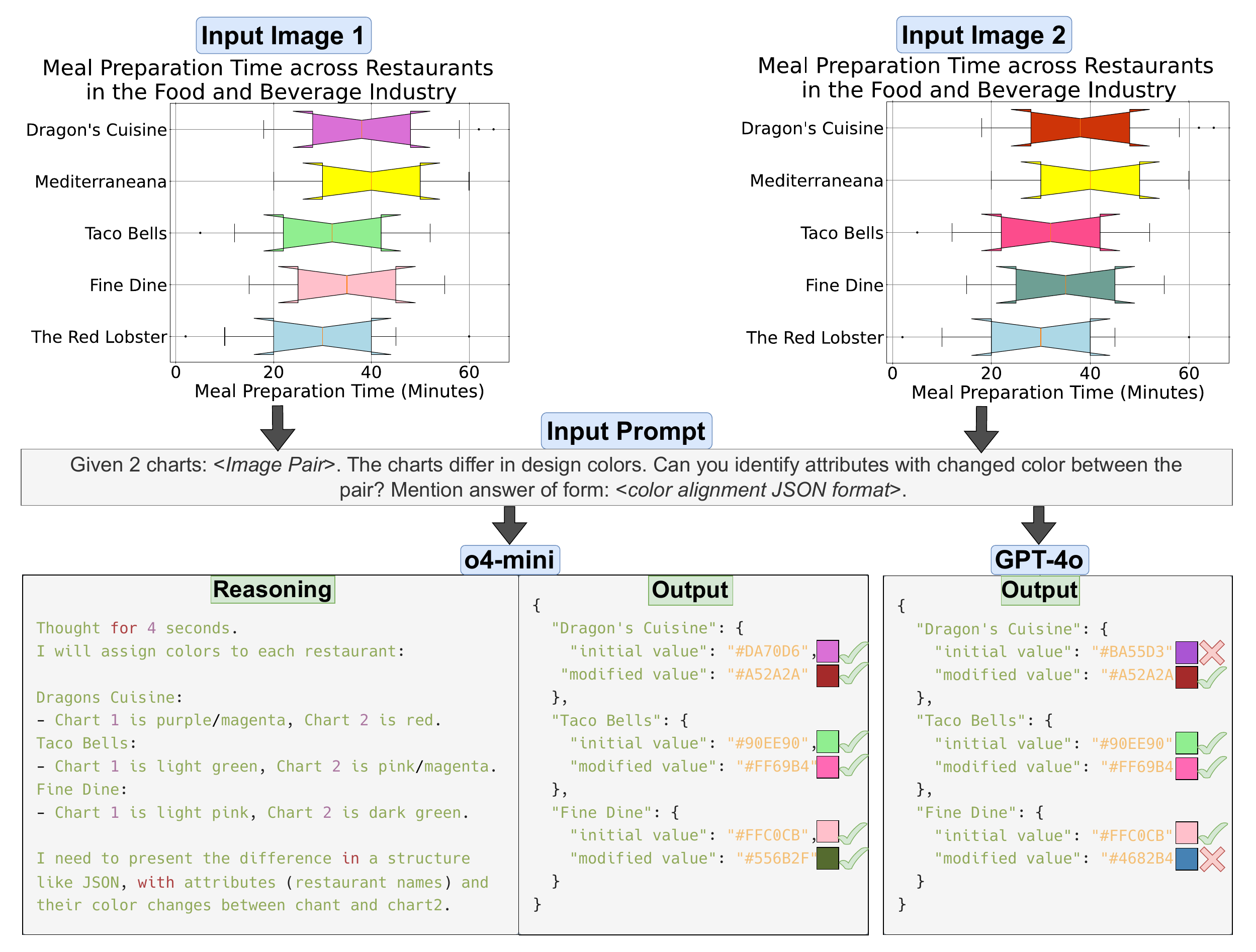}
   \vspace{-2em}
   % \caption{Multi-step approach of \textit{o4-mini} reasoning model on color-alignment evaluation example.}
   \caption{\textbf{Two-stage color alignment by o4-mini.}  
The o4-mini model automatically decomposes the task into a grounding step for the colors in each chart, followed by an output prediction of the alignment. This two-stage reasoning yields a more accurate result than GPT-4o, which performs alignment directly without intermediate grounding.}
    \label{fig:o4_mini_evaluation_example}
\end{wrapfigure}

% mark -> label -> value -> meaning
We propose a two-stage evaluation pipeline inspired by the multi-step approach of SOTA reasoning models, for example, color alignment by o4-mini \cite{openai2025o4mini} in Figure \ref{fig:o4_mini_evaluation_example}. The model's reasoning takes two steps: grounding the box colors in each chart, followed by dense alignment (comparison) of their grounded colors. This two-stage strategy performs complex, finer-level reasoning by ground-then-compare subtasks with efficient element-wise comparisons. It thus mitigates hallucinations and outperforms the one-stage strategy of GPT-4o, validating the importance of dense grounding for other tasks. 
% task decomposition approach and its ability for efficient multi-image dense alignment.

\begin{figure}[h]
\centering
\vspace{-2.em}
   \includegraphics[width=0.95\linewidth]{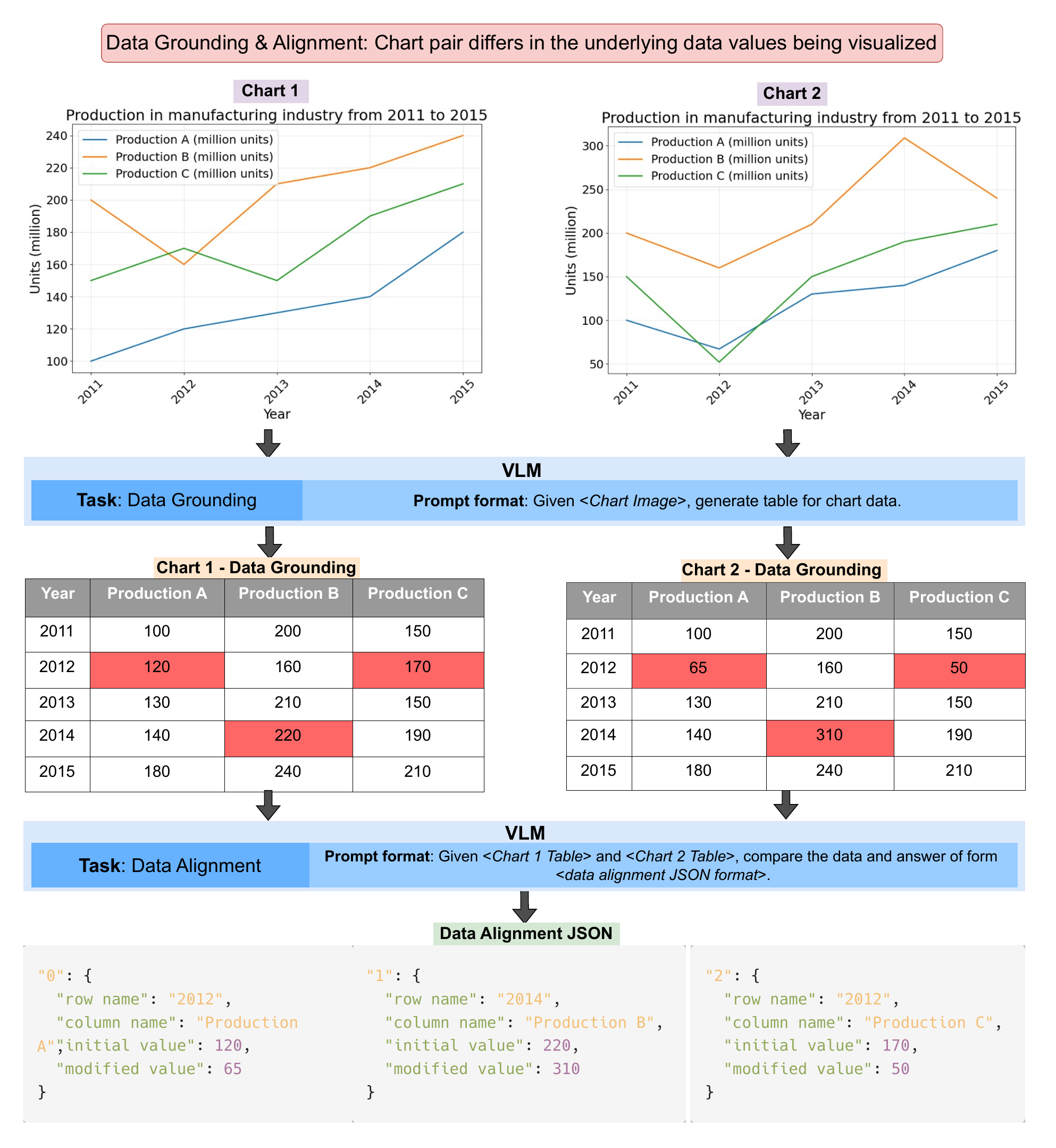}
   \vspace{-1em}
   \caption{\textbf{Two-Stage Evaluation Pipeline for \emph{Data Grounding \& Alignment} in \ours.} 
   % We formulate the two-stage approach with intermediate grounding for better comprehension of cross-modal information extraction and stronger dense level multi-image reasoning. 
   The first stage focuses on grounding the data visualized by each chart in a CSV table, while the second stage focuses on alignment, which aims to allocate the difference between the two tables and output a JSON file listing the different cells. The other two categories of tasks in \ours also adopt similar multi-stage pipelines, detailed in Figures \ref{fig:color_alignment_inference_pipeline}, \ref{fig:text_style_alignment_inference_pipeline}, \ref{fig:legend_alignment_inference_pipeline} of the Appendix. 
   % The first stage inference on chart images results in grounding to textual modality (chart data table here) allowing finer-level. 
   % This enables better reasoning toards finer-level alignment tasks driven by grounding information.
   }
\label{fig:data_alignment_inference_pipeline}
\vspace{-1.5em}    
\end{figure}

In our evaluation pipeline, the prompt in each stage consists of natural language instructions with a task-specific JSON template defining the output format. This enables better inswtruction following and flexible output parsing and evaluation. As shown in Figure~\ref{fig:data_alignment_inference_pipeline},   
The \textit{first-stage} performs grounding of data or certain attributes in the given charts. Such well-formatted element-wise representation facilitates subsequent dense alignment and QA tasks.  
% The interpretable nature and element-wise representation  subsequent reasoning for fine-grained alignment. 
The \textit{second-stage} compares the grounding results of the two charts from the first stage and produces a JSON file to list the dense alignment results. 
% involves VLM reasoning by applying discriminative comparison on the grounded results from first stage to perform the specific dense alignment task resulting in final JSON output.   

The second stage is critical to evaluating end-to-end alignment as it requires VLMs to perform semantic comparison over grounded outputs, beyond surface-level extraction. Compared to one-stage approaches, it mitigates grounding ambiguities and collects additional context, offering a more human-like assessment of alignment ability. More details of the pipeline are discussed in \ref{app:two_stage_pipeline}. \looseness-1

% In evaluation
% - Why 2nd stage perfect: (assertion) LLM performance on simple discriminative tasks
% - single stage: inconsistencies in CoT
% - stitched image: directly comparing finer-level differences difficult, seen - inconsistencies in chain-of-thought prompting

% \FloatBarrier (to avoid figure on separate page)

\subsection{Evaluation Metrics}
\textbf{Dense Grounding} performance is evaluated by the precision of the detected semantic elements in a given chart, e.g., values of visualized data, color of bars, legend position, font size. In the experiments, we report (1) \textit{Legend position} grounding's confusion matrix in Figure~\ref{fig:legend_grounding_error_analysis}; (2) Text-style grounding accuracy in Figure~\ref{fig:text_style_error_analysis}; (3) \textit{Color} grounding's L2 error of RGB values in Figure~\ref{fig:color_analysis_distance_wise}; and (4) \textit{Data} grounding performance in Figure~\ref{fig:grounding_impact_on_qa} is evaluated by the precision of predicted CSV using the SCRM metric introduced in StructChart \citep{xia2023structchart}. 
% \begin{equation}
% \mathcal{G}_\text{color} = \sqrt{(r - \hat{r})^2 + (g - \hat{g})^2 + (b - \hat{b})^2}
% \end{equation}

\textbf{Dense Alignment} performance is evaluated across four task categories: \textit{data alignment} (subtasks: 1-cell/2-cell/3-cell), \textit{color alignment}, \textit{text style alignment}, and \textit{legend alignment}. For each chart pair, the model is prompted to output a JSON file that lists the differences on possible attributes and their own values. The performance on the first three tasks is evaluated by a key-value alignment score, which assess the capability to identify the different elements (keys) between two charts and their associated values.  In contrast, 
% where we assess the model's ability to identify and localize specific chart elements that differ between image pairs. 
legend alignment score mainly focuses on comparing the different spatial positions of legends in two charts (values only) because the key (i.e., the position) is unique and fixed. More details of the keys and values are provided in Table~\ref{tab:alignment_tasks}, while the concrete definitions of the metrics are introduced in \ref{app:evaluation_metric_align}.

% on each chart pair $(x, x')$ has a JSON-formatted output that lists the differences across $N$ data cells or attributes, 
% , e.g., colors of bars/lines, altered data points, or text regions). 
% each associated with an alignment accuracy $\text{acc}_i(x, x')$ (whose detailed definition for each task is provided in \ref{app:evaluation_metric}). The alignment quality is measured by a score $s_\text{align}(x,x') = \nicefrac{1}{N} \sum_{i=1}^N\text{acc}_i(x, x')$, which is then normalized to $[0,10]$ and can be further averaged over multiple pairs. 

\textbf{Robustness} of data alignment to the variations of different visualization attributes, e.g., colors, legend positions, text style, is evaluated by the standard deviation of data alignment scores over multiple variations of the original chart pairs. We evaluate the robustness score under the variation of each attribute, and report the averaged scores over chart pairs. More details of the robustness score are provided in \ref{app:evaluation_metric_robust}. 

% and the \textbf{combined accuracy} is averaged and\textbf{ normalized }to $[0,10]$, resulting in the alignment score $\mathcal{S}_\text{align}$. This enables us to effectively differentiate model performance across tasks and quantify performance aspects for visualized data and visual attributes as part of dense alignment.The alignment score $\mathcal{S}_\text{align} = 10 \cdot \left( \frac{1}{N} \sum_{i=1}^N\text{acc}_i(x, x') \right)$. Task-wise metrics are discussed in \ref{app:evaluation_metric}

% \vspace{-1em}
% \begin{equation}
% \mathcal{S}_\text{align} = 10 \cdot \left( \frac{1}{N} \sum_{i=1}^{N}
% \text{acc}_i\left(\text{chart-pair}\right) \right)
% \end{equation}

\textbf{Grounding/Alignment affects QA Performance }To further analyze the impact of grounding/alignment quality on downstream QA tasks, we evaluate QA accuracy by following the protocols in 
% we adopt a binary correctness measure following its usage in corresponding QA tasks in 
ChartX~\citep{xia2024chartx}: string-based answers require an exact match, while numerical values are considered correct if they fall within a $5\%$ error margin; and investigate its correlation with the grounding/alignment performance. 
To this end, we adopt a two-stage QA that firstly extracts a CSV (table) file from a chart (data grounding), and then answers the question given the grounding result. We analyze how this two-stage QA's accuracy and its difference to the ordinary one-stage QA's accuracy vary with grounding/alignment quality, which results are reported in Figure~\ref{fig:qa_results}. 

\includecomment{
EVALUATION
FigureQA: Y/N
DVQA: numbers/%/label - exact match
ChartQA: csv file eval (compare value-by-value), annotators for Qs | numeric: 5% 
}

\section{Experiments \& Analysis}

We evaluated GPT-4o~\citep{hurst2024gpt} and four open-source VLM families: Phi-3.5 vision-instruct \citep{abdin2024phi}, InternVL-2.5 \citep{chen2024expanding}, LLaVA-1.6 \citep{liu2023improved}, QWEN-2.5 VL \citep{bai2025qwen2}. 
We also evaluated chart-specialized VLMs, including TinyChart~\citep{zhang2024tinychart} and ChartGemmap~\cite{masry2024chartgemma}. However, as discussed in Section~\ref{related_work:vlm_for_charts}, their task-specific training leads to a collapse of general instruction following capabilities and fails to output the JSON format required by evaluation. Further discussion and ablation study are provided in \ref{app:vlm_selection} and \ref{app:ablation_study}. \looseness-1

\begin{findingbox}[title={Finding 1}]
\label{par:model_weak_on_fine_grained_charts}
VLMs' dense grounding and alignment of data/color are not satisfying on complex charts.\looseness-1
\end{findingbox}
% \begin{figure*}[hbtp]
\begin{wrapfigure}[18]{r}{0.65\linewidth}
\centering
    \centering
    \vspace{-2.em}
    \includegraphics[width=0.95\linewidth]{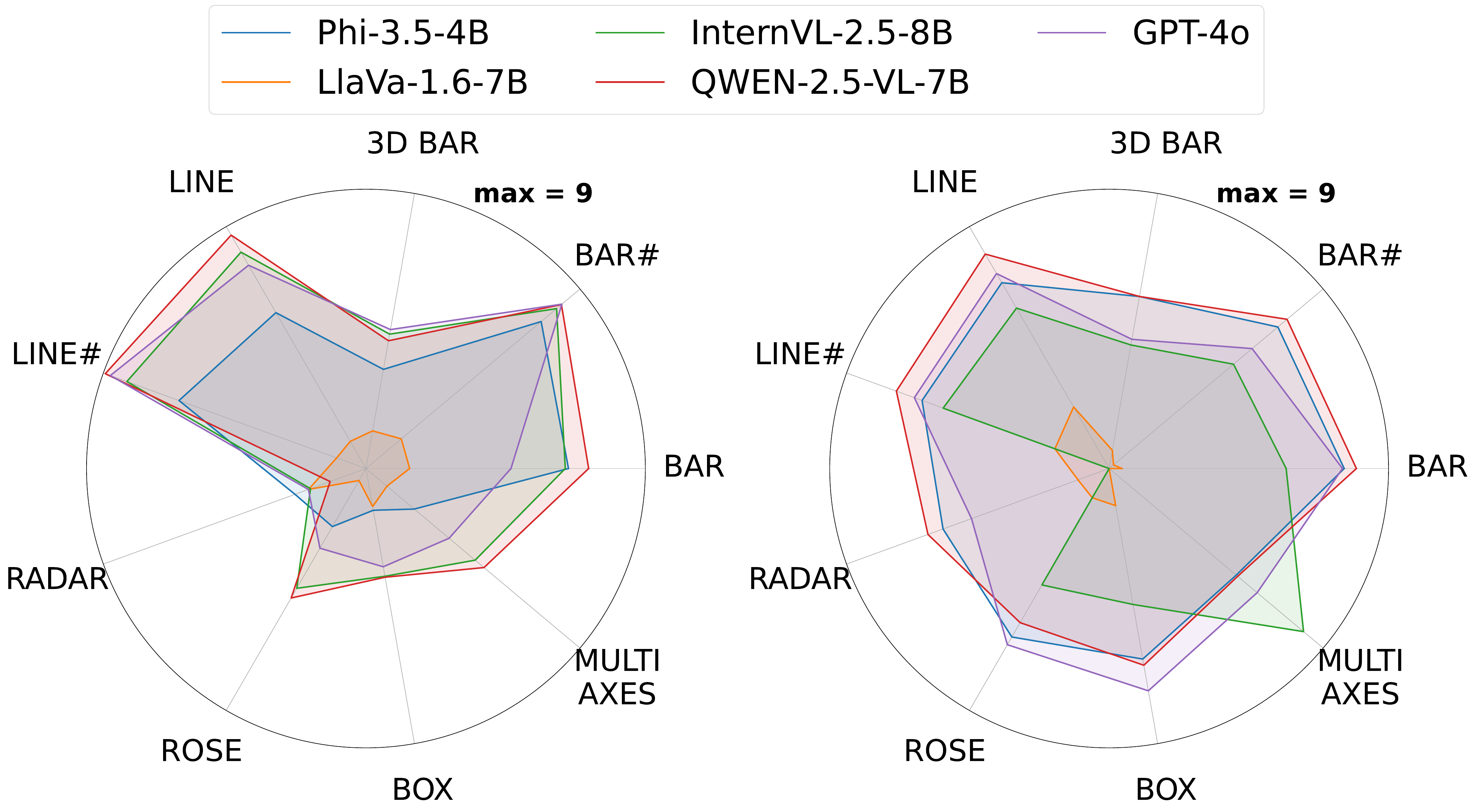}
    
    % \begin{subfigure}[t]{0.48\textwidth}
    %     \includegraphics[width=\linewidth]{imgs/eval_results_plots/model_comparison_1_cell_change.pdf}
    %     \label{fig:data_alignment_1_cell_model_comparison_radar_chart}
    % \end{subfigure}
    % \hfill
    % \begin{subfigure}[t]{0.48\textwidth}
    %     \includegraphics[width=\linewidth]{imgs/eval_results_plots/model_comparison_color_altered.pdf}
    %     \label{fig:color_alignment_radar_charts}
    % \end{subfigure}
    
    \vspace{-0.8em}
    \caption{\textbf{Left:} Comparing VLMs on \emph{Data Alignment} tasks on paried charts with \textbf{one-cell} difference. Llava-1.6 performs worse than most other VLMs, while QWEN-2.5-VL outperforms GPT-4o on most chart types. \textbf{Right: }\emph{Color alignment} on fine-grained visual elements (e.g., bars, lines, sectors) between two charts. Most VLMs perform better on simpler and more common charts, e.g., line/bar charts. Related discussion beneath Finding 1.}
    \label{fig:data_and_color_alignment_radar_plot}
    \vspace{-1em}
\end{wrapfigure}
% \end{figure*}
% \paragraph{P1: Model's weaker understanding of fine-grained chart constituents.} 
Compared to simpler and more common charts, e.g., bar/line charts and numbered bar/line charts, dense grounding/alignment on complex charts such as 3D/box/radar/rose/multi-axes charts with more components and irregular layouts is more challenging to most VLMs. 
% i.e. way visual encodings applied and plot attributes formulated. 
Despite the similar alignment performance for \textit{legend} (Figure~\ref{fig:legend_alignment_radar_charts}) and \textit{text-style} (Figure~\ref{fig:text_style_alignment_radar_charts}) between simple vs. complex charts, the \textit{color} and \textit{data} alignment (Figure~\ref{fig:data_and_color_alignment_radar_plot}) on complex charts are much poorer than those on simple charts. 
% weaker performance for fine-grained charts compared to the general ones. 
The color grounding requires identifying each constituent's visual encoding and corresponding color, while the data grounding needs to find the mapping from visual encoding to numeric values. Hence, complex layouts with more components make these tasks more difficult. In contrast, identifying the position of legends and text styles (which both have limited options) is easier and less affected by the chart complexity. 
% The model's inability to comprehend layout constituents in complex charts negatively impacts grounding for \textit{color} (it requires identifying visual encoding and their corresponding colors) \& \textit{data} (it requires mapping visual encoding to precise values) while components for \textit{legend} (legend's position) \& \textit{text-style} (text in title, legend, markings) alignment tend to be independent of the complexity of the charts. 

% \begin{figure*}[tp]
%     \centering
%     \vspace{-1em}
    
%     \begin{subfigure}[t]{0.48\textwidth}
%         \includegraphics[width=\linewidth]{imgs/eval_results_plots/model_comparison_legend_altered.pdf}
%         \caption{Legend Alignment}
%         \label{fig:legend_alignment_radar_charts}
%     \end{subfigure}
%     \hfill
%     \begin{subfigure}[t]{0.48\textwidth}
%         \includegraphics[width=\linewidth]{imgs/eval_results_plots/model_comparison_text_style_altered.pdf}
%         \caption{Text-Style Alignment}
%         \label{fig:text_style_alignment_radar_charts}
%     \end{subfigure}
    
%     \vspace{-0.5em}
%     \caption{(a) \textbf{Legend alignment} of legend positions. Phi-3.5 performs the worst while GPT-4o is best. Related discussion in Finding 1\&2. (b) \textbf{Text-style alignment} (size, weight, font). Worst: QWEN-2.5-VL, Best: GPT-4o. Discussion in Finding 1\&4.}
%     \label{fig:attr_alignment_comparison}
%     \vspace{-1em}
% \end{figure*}

% have similar level of correctness centered between Phi-3 \& GPT-4o.

% \paragraph{P4. Models perform differently on Text-Style}
% \label{par:models_perform_differently_on_text_style}
\begin{findingbox}[title={Finding 2}]
\label{par:models_perform_differently_on_text_style}
% Most VLMs suffer from biases when allocating the position of legends.\looseness-1
VLMs' text-style grounding and alignment performance is poor in general, and it varies across text size, weight, and font family. 
\end{findingbox}

\begin{wrapfigure}[11]{r}{0.40\linewidth}
    \vspace{-2em}
    \includegraphics[width=0.95\linewidth]{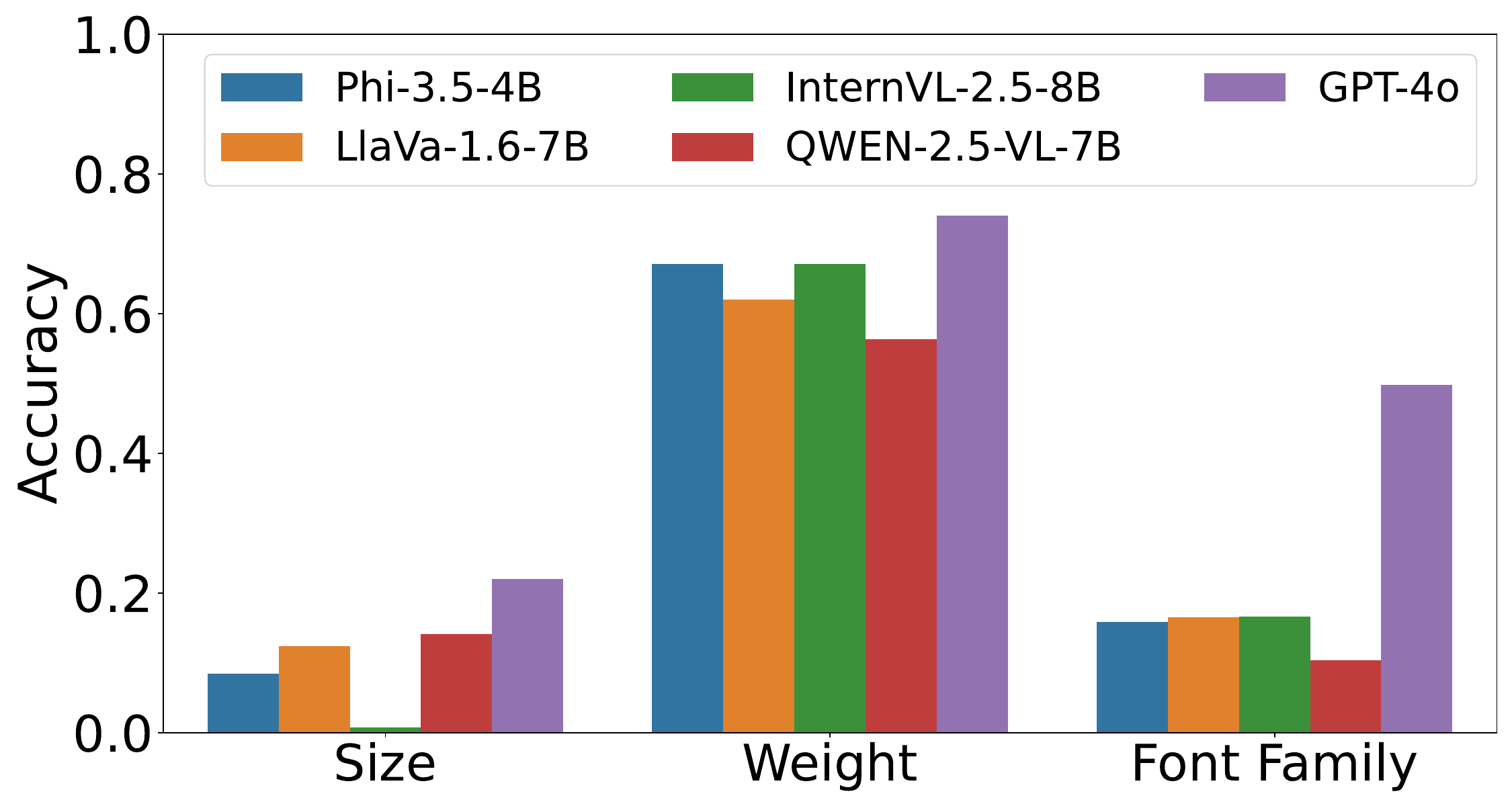}
    \vspace{-1em}
    \caption{\emph{Text-style grounding} on size, weight, and font family. The low accuracy of most VLMs highlights the lack of style knowledge (Finding 4).}
    \label{fig:text_style_error_analysis}
\end{wrapfigure}
% The text style alignment of models fails to show a common trend. 
% The alignment statistics (Fig. \ref{fig:text_style_error_analysis}) of characteristics: size, weight \& font family, for each the model shows differing performance, and overall weak understanding. 
Figure~\ref{fig:text_style_error_analysis} shows that most VLMs fail to detect the correct text size and font family, suffering from a $<$20\% accuracy (except GPT-4o's performance on font family grounding). These indicate a lack of knowledge on these two text attributes. VLMs' performance on text weight ((light/normal/bold)) is much better ($\sim$60\%) and close to each other, but still not satisfying. Although LLMs can select reasonable text sizes in code generation for plots, they tend to rely on the default sizes in their priors or relative sizes to other chart components. They still lack sufficient capability to identify exact text sizes in chart images.   
\begin{findingbox}[title={Finding 3}]
\label{par:weak_color_recognition_ability}
VLMs' weak color recognition ability.\looseness-1
\end{findingbox}

\begin{wrapfigure}[12]{r}{0.38\linewidth}
    \vspace{-2em}
    \includegraphics[width=0.95\linewidth]{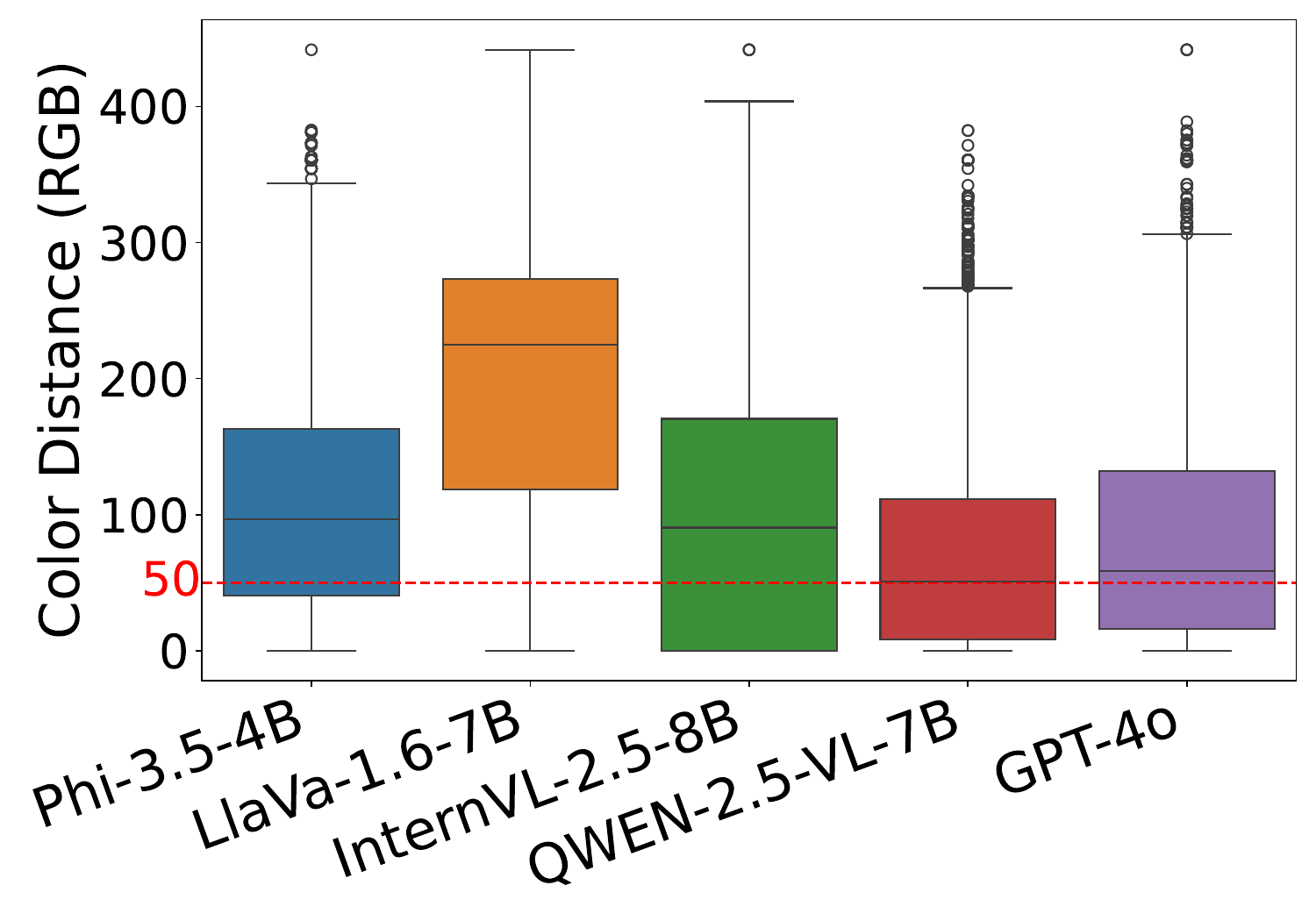}
     \vspace{-1em}
    \caption{\emph{Color grounding}'s L2 error in the RGB space, which median over VLMs $>$50 implies their weaknesses in color recognition (Finding 3). \looseness-1}
    \label{fig:color_analysis_distance_wise}
\end{wrapfigure}

As shown in Figure~\ref {fig:color_analysis_distance_wise}, all models' color grounding error (L2 distance in RGB space) has a median exceeding $50$. This implies their inability to understand color shades beyond common ones, e.g., red, blue, green, etc., which exposes their weaknesses in color recognition.  
% VLMs have weak understanding of color recognition. The median distance(RGB) for color recognition exceeds 50 for all models. This suggests VLM's inability to understand color shades beyond broad colors, e.g., red, blue, green, etc. 

% \begin{wrapfigure}[14]{r}{0.4\linewidth}
%     \includegraphics[width=0.95\linewidth]{imgs/color_recognition.pdf}
%     \caption{\textbf{Color recognition} measured by L2 errors in RGB space. Median errors exceed 50 for all VLMs, indicating weak color recognition (Finding 3).}
%     \label{fig:color_analysis_distance_wise}
% \end{wrapfigure}

The lack of color understanding affects the perception of detailed differences in charts and leads to misalignment in color-conditioned reasoning tasks. Consequently, the VLMs' performance in color alignment tasks (Figure \ref{fig:data_and_color_alignment_radar_plot}) is consistent with that on color grounding. 
% The Color-aware training could be applied for specific classes of tasks in charts, art \& design, and medical imaging by focusing on color patches in the images. 
These results suggest to improve the color understanding capability by adding more color-sensitive data to VLM training. \looseness-1

\begin{figure*}[bhtp]
\centering
\includegraphics[width=0.92\linewidth]{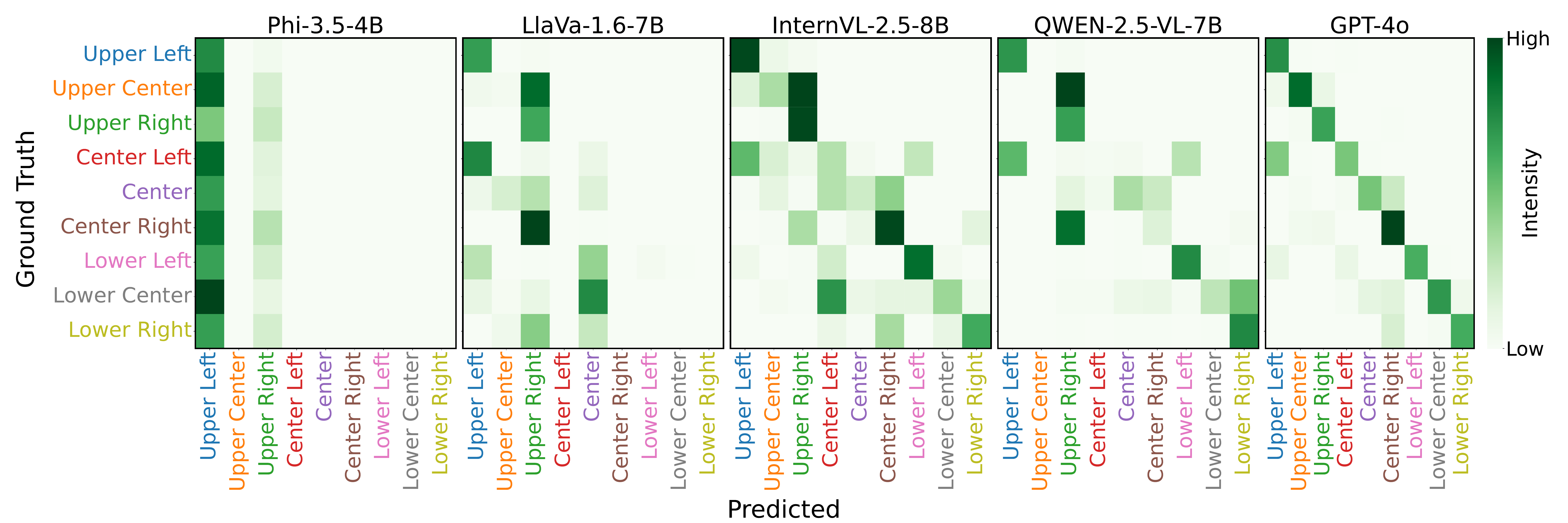}
\vspace{-1em}
   \caption{Confusion matrix of \emph{legend position grounding}. The dark non-diagonal entries show the fail patterns and imply the biases of incorrectly identifying position-$i$ as position-$j$. 
   % confusion-matrix from the analysis of legend-alignment predictions showcases biases for models in predicting legend positions. 
   Phi-3.5 exhibits a severe bias towards the \textit{upper-left} position while GPT-4o shows the minimal bias. More discussion is provided below Finding 2.
   % \S\ref{par: model_bias_in_legend_prediction}. 
   \looseness-1
   } 
    \label{fig:legend_grounding_error_analysis}
    \vspace{-1em}
\end{figure*}

% \paragraph{P2. Model's bias in legend prediction.}
\begin{findingbox}[title={Finding 4}]
\label{par: model_bias_in_legend_prediction}
Spatial reasoning bias: Most VLMs suffer from biases when allocating the position of legends.\looseness-1
\end{findingbox}

The grounding of the legend's position (Figure~\ref {fig:legend_grounding_error_analysis}) suffers from a strong bias of pretrained VLMs. 
% As the grounding information extracted directs the flow of the alignment, generation bias due to model prior impacts alignment: as seen in legend alignment, the grounding analysis (Fig. \ref{fig:legend_grounding_error_analysis}) indicates incidence of strong priors which bias the prediction towards corresponding positions. 
The Phi-3.5 model shows the strongest prior towards the \textit{upper-left} position. 
The 7-8B scale VLMs, e.g., LlaVa-1.6, Inten-VL-2.5, QWEN-2.5-VL, all show a similar level of bias but towards the \textit{upper-right} position instead. 
The GPT-4o model exhibits the minimal bias among all evaluated VLMs.
% and hence performing good grounding of the legend. 
The grounding bias strongly affects the legend alignment (Figure ~\ref{fig:legend_alignment_radar_charts}) where Phi-3.5 performs the worst, GPT-4o has the best performance, while the other 3 models' performance is between them. 

\begin{figure*}[htb]
    \centering
    \vspace{-1em}
    
    \begin{subfigure}[t]{0.48\textwidth}
        \includegraphics[width=\linewidth]{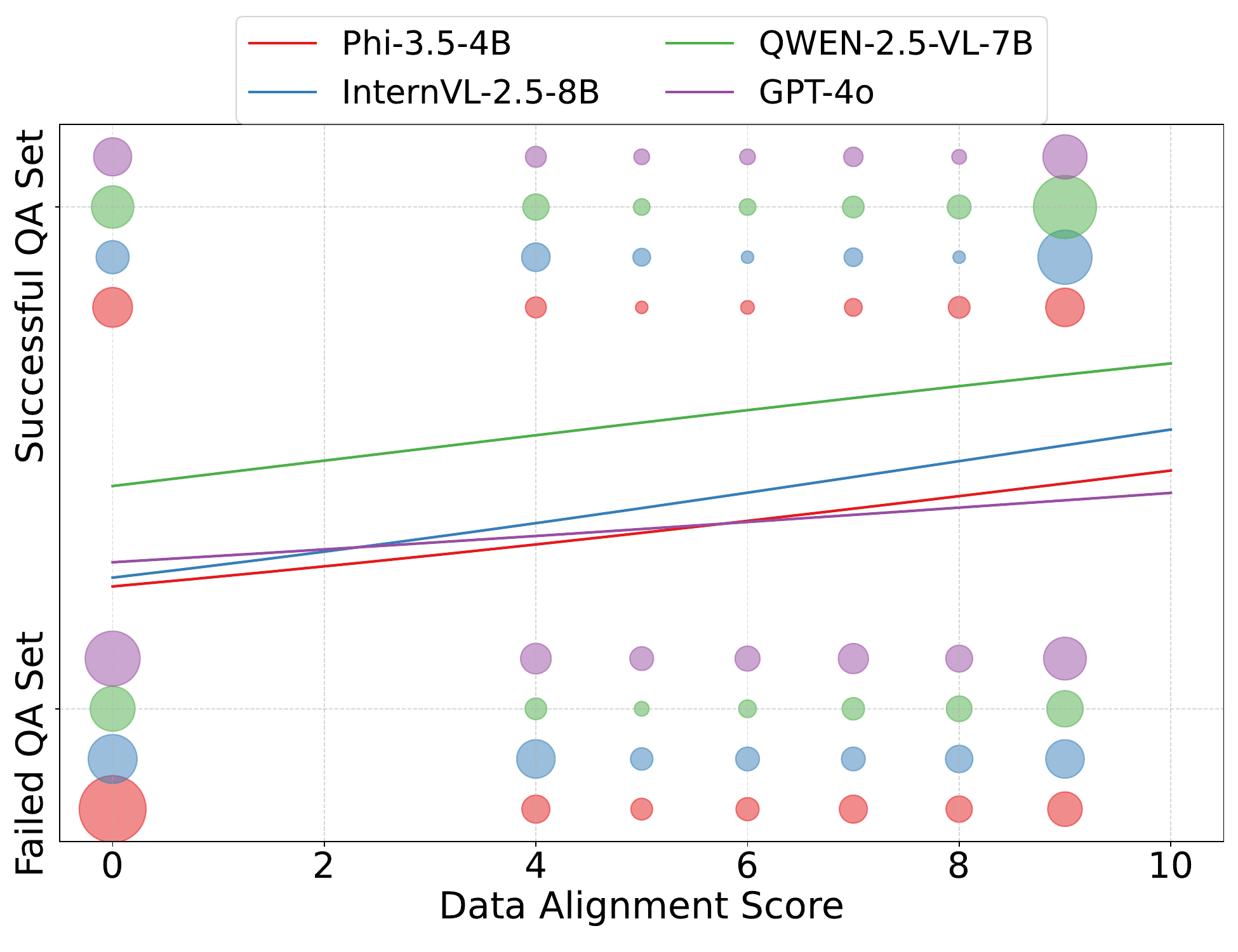}
        \caption{Data Alignment correlates with QA performance.}
        % \caption{Regression of QA accuracy on visualized-data alignment score}
        \label{fig:regression_of_qa_on_alignment}
    \end{subfigure}
    \hfill
     \begin{subfigure}[t]{0.48\textwidth}
        \includegraphics[width=\linewidth]{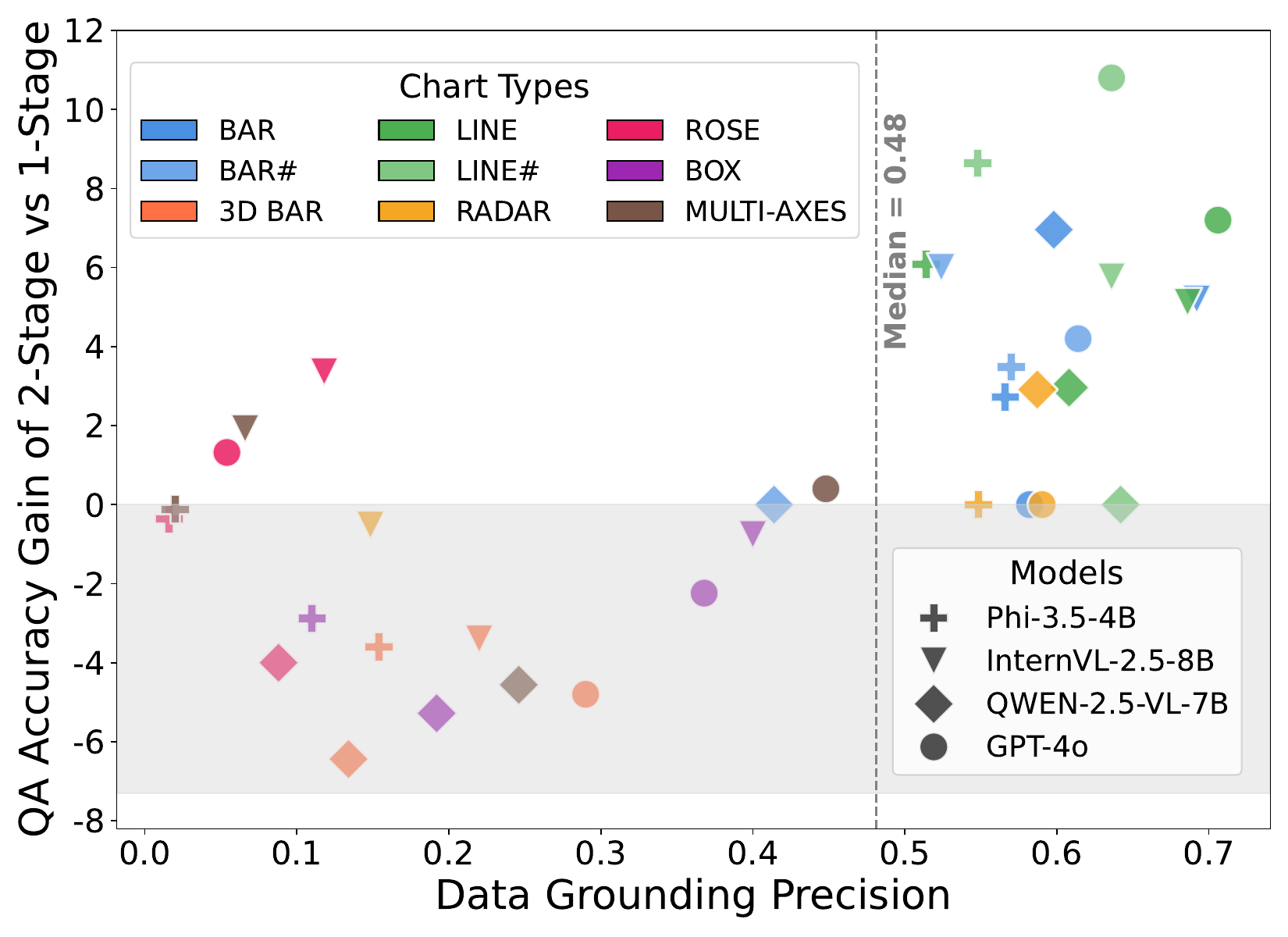}
        \caption{Data Grounding's impact on QA Performance.}
        \label{fig:grounding_impact_on_qa}
    \end{subfigure}
    
    \vspace{-0.5em}
    \caption{\textbf{(a)} 
    % QA correctness (0/1) for each chart is regressed on the data alignment score (0 - 10) of its paired perturbation.  
    shows that the failed (successful) QA tasks decrease (increase) with the data alignment score, underscoring the importance of data alignment capability of VLMs on downstream chart reasoning tasks.
    \textbf{(b)} shows that precise (poor) data grounding leads to positive (negative) gain on QA tasks, indicating the importance of data grounding on downstram tasks. 
    % shows charts with above-median grounding precision consistently yield a positive weighted $\Delta$QA, demonstrating that stronger grounding directly boosts downstream QA. 
    More discussion can be found beneath Finding 6.}
    \label{fig:qa_results}
    \vspace{-1em}
\end{figure*}

\begin{findingbox}[title={Finding 5}]
\label{par:downstream_qa}
Poor (precise) grounding and alignment degrade (improve) downstream QA performance.
\end{findingbox}

Figure~\ref{fig:grounding_impact_on_qa} demonstrates that precise (poor) grounding of chart-visualized data boosts (degrades) QA performance. It validates grounding as a gateway to extract structured data from charts for reliable downstream reasoning. Notably, the greatest gains are achieved on simple chart types (bar/line charts and numbered bar/line charts) due to better numeric understanding of these charts' visualized data, as discussed in Finding 1.  
Figure~\ref{fig:regression_of_qa_on_alignment} shows a steady rise of QA accuracy (predicted) with the data alignment score, demonstrating the importance of dense chart understanding to QA reasoning.
These findings position grounding and alignment as essential prerequisites for chart reasoning.

\begin{findingbox}[title={Finding 6}]
\label{par:presence_of_scaling_law}
% VLMs' weak scaling law on chart grounding and alignment tasks. 
VLMs follow the scaling law on most dense alignment tasks.
\end{findingbox}

\begin{wrapfigure}[8]{r}{0.5\linewidth}
\vspace{-6.5em}
   \includegraphics[width=0.95\linewidth]{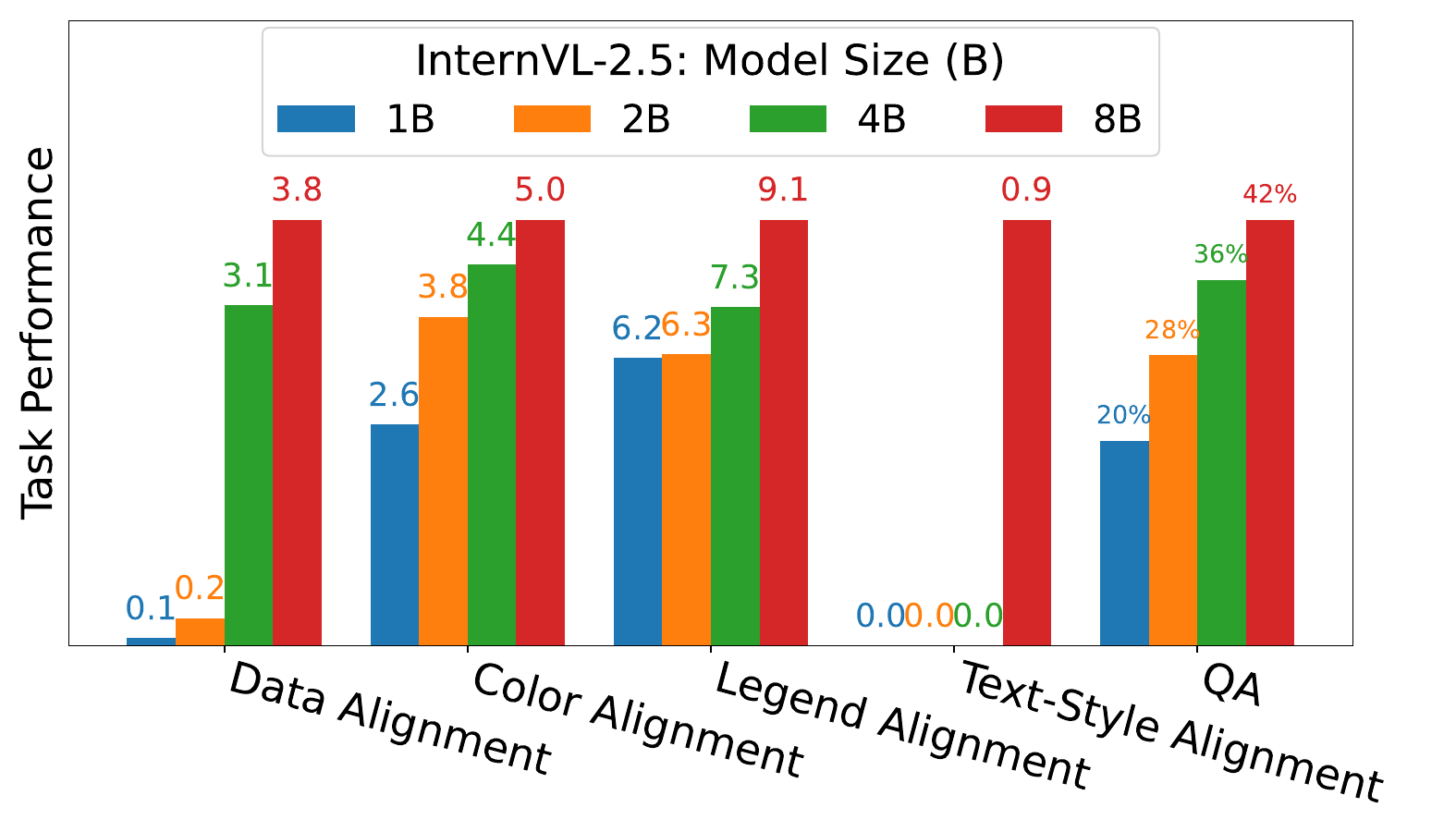}
   \vspace{-1.5em}
   \caption{Alignment performance of VLMs with different sizes from the InternVL-2.5 family. Results of other VLMs are reported in Appendix~\ref{fig:scaling_law_qwen_and_llava}.}

    \label{fig:scaling_law_plot}
    \vspace{-1em}
\end{wrapfigure}

As shown in Figure \ref{fig:scaling_law_plot}, we observed a consistent scaling law across most dense alignment subtasks, except for Text-Style Alignment. The deviation arises from the relatively greater complexity of the JSON template in this task, which led to a significantly higher number of failures where InternVL-2.5 produced incorrect JSON formats. 

\section{Conclusion}
We introduce \ours, the first benchmark for fine-grained chart grounding and multi-chart dense alignment in vision–language models (VLMs).
Our evaluations across diverse chart types reveal persistent challenges, including perceptual bias, weak attribute understanding, and limited spatial reasoning especially on complex visual representations.
Experiments with our novel two-stage pipeline show effectiveness of intermediate grounding in improving dense alignment, and the impact of grounding and alignment accuracy for enhance downstream question answering, establishing these capabilities as essential foundations for robust chart understanding.

\bibliography{iclr2026_conference}
\bibliographystyle{iclr2026_conference}

\clearpage
\appendix
\section{Appendix}

\subsection{LLM usage statement}

LLMs were used in the work as general purpose writing aid (e.g. to polish grammar and phrasing) and to assist with literature search. All substantive research ideation, experiments and analysis has been conducted by the authors. 

\subsection{Limitations}

Our work focuses on VLM evaluations and do not assess model fine-tuning. While such approaches might yield stronger results, they diverge from our goal of studying general purpose VLMs for dense level understanding. For dataset construction despite availability of chart datasets with more sophisticated real-world chart examples, we selected the ChartX \cite{xia2024chartx} dataset because it provides precise chart information in form of csv data and plotting code which is essential for generating precise ground truth values for the evaluation of dense grounding and alignment.
% Our work has the following limitations:-
% \begin{itemize}
%     \item Model Training: We focus on zero-shot evaluations for our work, and don't assess few-shot or instruction tuned performance. They may yield better performance but deflect from the problem statement of general purpose VLM's dense-level understanding. 
%     \item Real-World Chart Corpus: Various datasets contain more sophisticated real-world examples. However due to requirement of precise ground-truth for dense-alignment evaluation, we chose the ChartX dataset due to availability of plotting-code and corresponding csv data. 
% \end{itemize}

\subsection{Dataset Construction}
\label{app:dataset_curation}

\begin{algorithm}[htbp]
\caption{\ours Dataset Construction: \textit{Data Grounding and Alignment} Subset}
\label{algo:data}
\KwIn{
  Source dataset $\mathbf{D}_{\text{ChartX}}=\{(T_i,S_i)\}_{i=1}^{N}$ from ChartX~\citep{xia2024chartx}, where $T_i$ is a CSV table and $S_i$ is the corresponding plotting script, $N$ is number of instances;
  Number of cells to modify $k \in \{1,2,3\}$;
  Scaling range $[\alpha_{\min}, \alpha_{\max}]$.
}
\KwOut{%
    % Constructed dataset $\mathbf{D}^{(data)}_{\ours} = \{(x_i, x'_i, gr_i, al_i)\}_{i=1}^{M}$,\\
    % where $x_i$ and $x'_i$ are chart image pairs, $gr_i$ is the grounding label (CSV table), and $al_i$ is the alignment label.
    Constructed dataset $\mathbf{D}^{(\mathrm{data})}_{\ours}=\{(x_i,x'_i,y^g_i,y^a_i)\}_{i=1}^{M}$, where $x_i,x'_i$ are chart images, $y^g_i$ is the grounding label, and $y^a_i$ is the alignment label, $M$ is number of instances.
}

\ForEach{$(T, S) \in \mathbf{D}_{\text{ChartX}}$}{
    Parse table $T$ to obtain a set of all cells $C = \{(r, c, v_{r, c})\}$, where $r$ and $c$ denote cell's row label and  column label respectively, and $v_{r,c}$ the corresponding cell value\;
    
    Identify candidate cells $C' \subseteq C$ with unique values\;
    \If{$|C'| < k$}{\textbf{skip} this chart;}
    
    Sample $k$ cells $\{(r_i, c_i, v_{r_i, c_i})\}_{i=1}^{k}$ from $C'$\;
    Sample scaling factors $\{\alpha_i\}_{i=1}^{k}$ from scaling range $[\alpha_{\min}, \alpha_{\max}]$\;
    
    Initialize $T' \leftarrow T$ and $S' \leftarrow S$\;
    \ForEach{$(r, c, v_{r, c}) \in C'$}{
        Compute modified value $v'_{r,c} = \alpha_i \cdot \mu_{c}$, where $\mu_{c}$ is the mean of cells in column $c$\;
        
        % \If \neg{unique match of $v_{r_i,c_i}$ in $S$}{
        %     \textbf{skip current example and continue to next}\;
        % }
        \If{not (unique match of $v_{r,c}$ in $S$)}{
           \textbf{skip} this chart\;
        }
        
        Replace $v_{r,c}$ with $v'_{r,c}$ in $T'$ and $S'$ \;
    }
    
    Execute $S$ and $S'$ to generate chart images $x$ and $x'$\;
    \If{$x$ and $x'$ generation succeed}{
        Create instance $(x, x', y^g, y^a)$ where $y^g=(T, T')$ and $y^a=\{(r_i, c_i, v_{r_i,c_i}, v'_{r_i,c_i})\}_{i=1}^{k}$\;
        Append $(x, x', y^g, y^a)$ to $\mathbf{D}^{(data)}_{\ours}$\;
    }
}
\end{algorithm}
% \begin{algorithm}[htbp]
% \caption{\ours dataset construction: \textit{Data Grounding \& Alignment}}
% \label{alg:cell_edit}
% \KwIn{%
%     ChartX \cite{xia2024chartx} dataset $\mathcal{D}$ with  CSV table and Python plotting script;\\
%     Number of data points to modify $k$;\\
% }
% \KwOut{%
%     (1) Chart image-pairs, (2) Ground truth - grounding labels: csv-table for each chart, alignment labels: describe difference between image-pairs on data points;
% }
% \ForEach{chart instance: given csv-table and plotting-script}{
%     Parse the (ground truth) csv-table $T$ of $x$\;
%     Identify candidate cells with unique values\;
%     \If{fewer than $k$ candidate cells}{\textbf{continue}}
%     Randomly choose $k$ candidate cells and random scaling factors\;
%     \ForEach{chosen cell $(r_i, c_i)$}{
%         Compute a modified value by \textit{scaling the column mean}\;
%         Record original and modified values (for label generation)\;
%     }
%     Modify the Python plotting script of $x$ by replacing each original value
%     with its modified value (only if a unique match exists)\;
%     Execute the scripts: initial and modified, to generate the image-pair\;
%     \If{execution succeeds}{
%         Add to ground truth label file: grounding label (csv table) and alignment label (JSON: cell change, each described by [row name, column name, initial value, modified value])\;
%     }
% }
% \end{algorithm}

\begin{algorithm}[htb]
\caption{\ours Dataset Construction: \textit{Attribute Grounding and Alignment} Subset}
\label{algo:attr}
\KwIn{%
    Source dataset $\mathbf{D}_{\text{ChartX}} = \{S_i\}_{i=1}^{N}$ from ChartX~\citep{xia2024chartx}, where $S_i$ is the plotting script, $N$ is number of instances; Set of attribute types $B = \{\text{color}, \text{legend}, \text{text style}\}$.
}
\KwOut{%
    Constructed dataset $\mathbf{D}_{\ours}^{\text{(attribute)}} = \{(x_i, x'_i, b_i, y^{g}_i, y^{a}_i)\}_{i=1}^{M}$, where
    $x_i$, $x'_i$ are chart images,
    $b_i \in B$ is the attribute type,
    $y^{g}_i$ is the grounding label, 
    $y^{a}_i$ is the alignment label,
    $M$ is number of instances.
}
\ForEach{$(T, S) \in \mathbf{D}_{\text{ChartX}}$}{
    Parse plotting script $S$ using regex to detect plot attributes\;
    
    $\text{color\_list} \leftarrow$ locate \textit{unique} color array in $S$, corresponding to visual encodings (e.g., bars/lines/boxes)\;
    
    $\text{legend\_position} \leftarrow$ extract position parameter from \texttt{legend(..., loc=$\cdot$)} in $S$\;
    
    $\text{text\_style} \leftarrow$ parse \texttt{rcParams} for size, weight, and font family for regions (\textit{title, legend, axes labels, axes ticks})\;
    
    Collect detected attributes $\{\text{color\_list}, \text{legend\_position}, \text{text\_style}\}$\;
    
    \If{any attribute value is undefined or ambiguous}{
        \textbf{skip this chart}\;
    }
        
    \tcp{Generate modified versions for each attribute type}
    \ForEach{attribute type $b \in B$}{
        Initialize $S' \leftarrow S$, $y^g \leftarrow \emptyset$, and $y^a \leftarrow \emptyset$\;
        
        \uIf{$b = \text{color}$}{
            Sample new color list $\text{color\_list}'$ by randomly replacing a subset of colors\;
            Replace color array in $S'$ with $\text{color\_list}'$\;
            $y^g \leftarrow (\text{color\_list}, \text{color\_list}')$\;
            $\text{changed\_colors} \leftarrow \{(c_{\text{old}}, c_{\text{new}}) \mid c_{\text{old}} \neq c_{\text{new}}\}$\;
            $y^a \leftarrow \{$``type'': ``color'', ``changed'': $\text{changed\_colors}$\}\;
        }
        \uElseIf{$b = \text{legend}$}{
            Sample new legend position $\text{legend\_position}' \in \{\text{`upper left'}, \text{`upper right'}, \dots\}$\;
            Replace \texttt{loc} parameter in $S'$ with $\text{legend\_position}'$\;
            $y^g \leftarrow (\text{legend\_position}, \text{legend\_position}')$\;
            $y^a \leftarrow \{$``type'': ``legend'', ``changed'': $\text{legend\_position}'$\}\;
        }
        \uElseIf{$b = \text{text style}$}{
            Sample new text style parameters $\text{text\_style}'$ (font size, weight, or family)\;
            Update \texttt{rcParams} in $S'$ with $\text{text\_style}'$\;
            $y^g \leftarrow (\text{text\_style}, \text{text\_style}')$\;
            $\text{changed\_fields} \leftarrow \{(k, v_{\text{old}}, v_{\text{new}}) \mid \text{text\_style}[k] \neq \text{text\_style}'[k]\}$\;
            $y^a \leftarrow \{$``type'': ``text style'', ``changed'': $\text{changed\_fields}$\}\;
        }
        
        Execute $S'$ to generate modified chart image $x'$\;
        \If{$x'$ generation succeeds}{
            Create instance $(x, x', b, y^{g}, y^{a})$\;
            Append $(x, x', b, y^{g}, y^{a})$ to $\mathbf{D}_{\ours}^{\text{(attribute)}}$\;
        }
    }
}
\end{algorithm}

\begin{algorithm}[hbt]
\caption{\ours Dataset Construction: \textit{Robustness Set Generation}}
\label{algo:robustness}
\KwIn{%
    Source dataset $\mathbf{D}_{\text{ChartX}} = \{(T_i, S_i)\}_{i=1}^{N}$,
    where $T_i$ is a CSV table and $S_i$ is the corresponding plotting script, $N$ is number of instances;
    Number of cells to modify $k \in \{1,2,3\}$;
    Scaling range $[\alpha_{\min}, \alpha_{\max}]$;
    Visual variations per instance $d = 5$;
    Set of attribute types: $B = \{\text{color}, \text{legend}, \text{text style}\}$.
}
\KwOut{%
    $\mathbf{D}_{\ours}^{\text{(robust)}} = \{\{(x^{(j)}_{i}, x'^{(j)}_{i})\}_{j=1}^{d}, y^g_i, y^a_i, at_i\}_{i = 1}^{M}$
    where
    $x^{(j)}_i$, $x^{'(j)}_i$ are chart images for variation $j$,
    $b_i \in B$ is the attribute type being varied,
    $y^{g}_i$ is the grounding label, 
    $y^{a}_i$ is the alignment label.
}

\ForEach{$b \in B$}{
    \ForEach{$(T, S) \in \mathbf{D}_{\text{ChartX}}$}{
        \tcp{Apply data modification (Algorithm~\ref{algo:data})}
        Parse $T$ to extract cells $\{(r, c, v_{r,c})\}$\; identify unique-value cells $C'$ \textbf{if} $|C'| >=  k$\;
        % \If{$|C'| < k$ \textbf{or} attribute $b$ undefined in $S$}{\textbf{skip this chart}}
        
        Sample $k$ cells $\{(r_i, c_i)\}_{i=1}^{k}$ from $C'$ and scaling factors $\{\alpha_i\}_{i=1}^{k}$ from $[\alpha_{\min}, \alpha_{\max}]$\;
        Create modified table $T'$ and script $S'$ by replacing $v_{r_i,c_i}$ with $v'_{r_i,c_i} = \alpha_i \cdot \mu_{c_i}$\;
        \If{any $v_{r_i,c_i}$ has non-unique match in $S$}{\textbf{skip} this chart}
        
        Set $y^g \leftarrow (T, T')$ and $y^a \leftarrow \{(r_i, c_i, v_{r_i,c_i}, v'_{r_i,c_i})\}_{i=1}^{k}$\;
        
        \tcp{Generate base pair and visual variations}
        Execute $S$ and $S'$ to generate base charts $x^{(0)}$ and $x'^{(0)}$\;
        \If{generation fails}{\textbf{skip this chart}}
        
        Initialize $\mathcal{P} \leftarrow \emptyset$\;
        \For{$j = 1$ \KwTo $v$}{
            Sample variation $\Delta_j$ for attribute $b$ (color/legend/text style)\;
            Apply $\Delta_j$ to both $S$ and $S'$ to create $S_j$ and $S'_j$\;
            Execute $S_j$ and $S'_j$ to generate $x^{(j)}$ and $x'^{(j)}$\;
            \If{generation succeeds}{Add $(x^{(j)}, x'^{(j)})$ to $\mathcal{P}$}
        }
        
        \If{$|\mathcal{P}| = v$}{
            Append $\{\{(x^{(j)}, x'^{(j)})\}_{j=1}^{d}, y^g, y^a, at\}$ to $\mathbf{D}_{\ours}^{\text{(robust)}}$\;
        }
    }
}
\end{algorithm}

We used ChartX dataset \cite{xia2024chartx} as source dataset for our ChartAlignBench curation. ChartX contains plotting-code and csv data-table for the chart with extremely high level of precision thus offering the flexibility for performing finer-level changes along with ground-truth generation capabilities. It contains diverse chart types of varying complexities, and chart data from multiple domains. Hence enabling analysis across charts of varying difficulties.
% \subsubsection{Source: ChartX dataset}
% - large corpus, code, various chart types (simple, complex), diverse categories
% We use the ChartX dataset \cite{xia2024chartx} as source for curating our ChartAlign (Fig. ) dataset. 
% \begin{itemize}
%     \item Precise modalities: ChartX dataset contains chart images, plotting code, csv data table with extremely high level of precision due to sequential csv \& code to image generation approach. Thus enabling fulfillment our dataset generation routine's requirement of chart data \& plot attribute alteration on modalities.  
%     \item Chart types: ChartX dataset encompasses diverse chart types including general ones (like bar, line), along with specific ones (like box, radar, rose) making it suitable source for required comparative analysis for different chart types of differing structures and varying difficulties. 
%     \item Chart topics: ChartX dataset includes charts from 22 different topics covering all aspects of various domains including commerce, industry, society, culture, and lifestyle. Hence making evaluation robust against topic-specific semantic understanding of models' impact on dense alignment. 
% \end{itemize}

% \subsubsection{ChartAlignBench}
We utilize \textit{perturbations} for generating fine-grained variations for given chart thus helping build dense-alignment pairs. Chart's plotting-code is perturbed for precise data or attribute changes based on rigorous formatting check using regex-based search and replace, resulting in chart image generation from code execution. 

The csv availability and attribute information enable accurate ground-truth generation. Generated pairs for data alignment and attribute alignment include randomly assigned changes, and robustness sets include diverse attribute values for meticulous and unbiased evaluation.

The algorithmic description for generating chart pairs for \textit{Data Grounding \& Alignment} \ref{algo:data}, \textit{Attribute Grounding \& Alignment}  \ref{algo:attr}, \textit{Robustness} \ref{algo:robustness} describe the process in detail.

\subsection{A Two-Stage Evaluation Pipeline Details}
\label{app:two_stage_pipeline}

We utilize natural-language based instructions for zero-shot inference to enable simple execution with minimal task specific nuances for strong generalization across various models. 

\begin{figure*}[h]
\centering
\vspace{-1em}
   \includegraphics[width=0.85\linewidth]{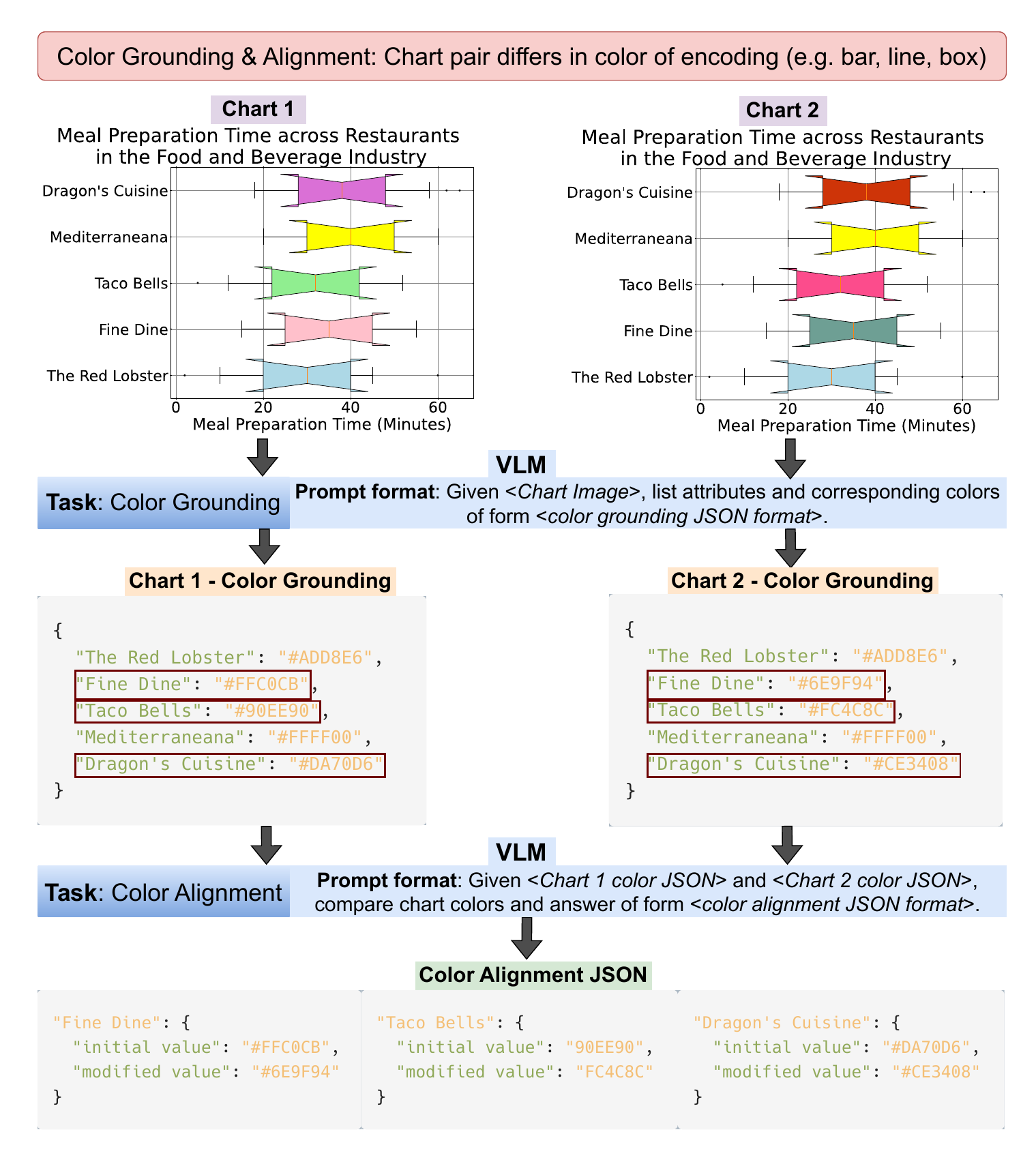}
   \vspace{-1em}
   \caption{\textbf{Two-Stage Evaluation Pipeline for \emph{Color Grounding \& Alignment} in \ours.} 
   The first stage focuses on grounding the color for visual encodings in each chart, while the second stage focuses on alignment, which aims to evaluate the colors for visual encodings and output a JSON file listing the visual encodings which differ in color values between the chart pair.}
\label{fig:color_alignment_inference_pipeline}
\vspace{-1em}    
\end{figure*}

\begin{figure*}[htp]
\centering
\vspace{-1em}
   \includegraphics[width=0.85\linewidth]{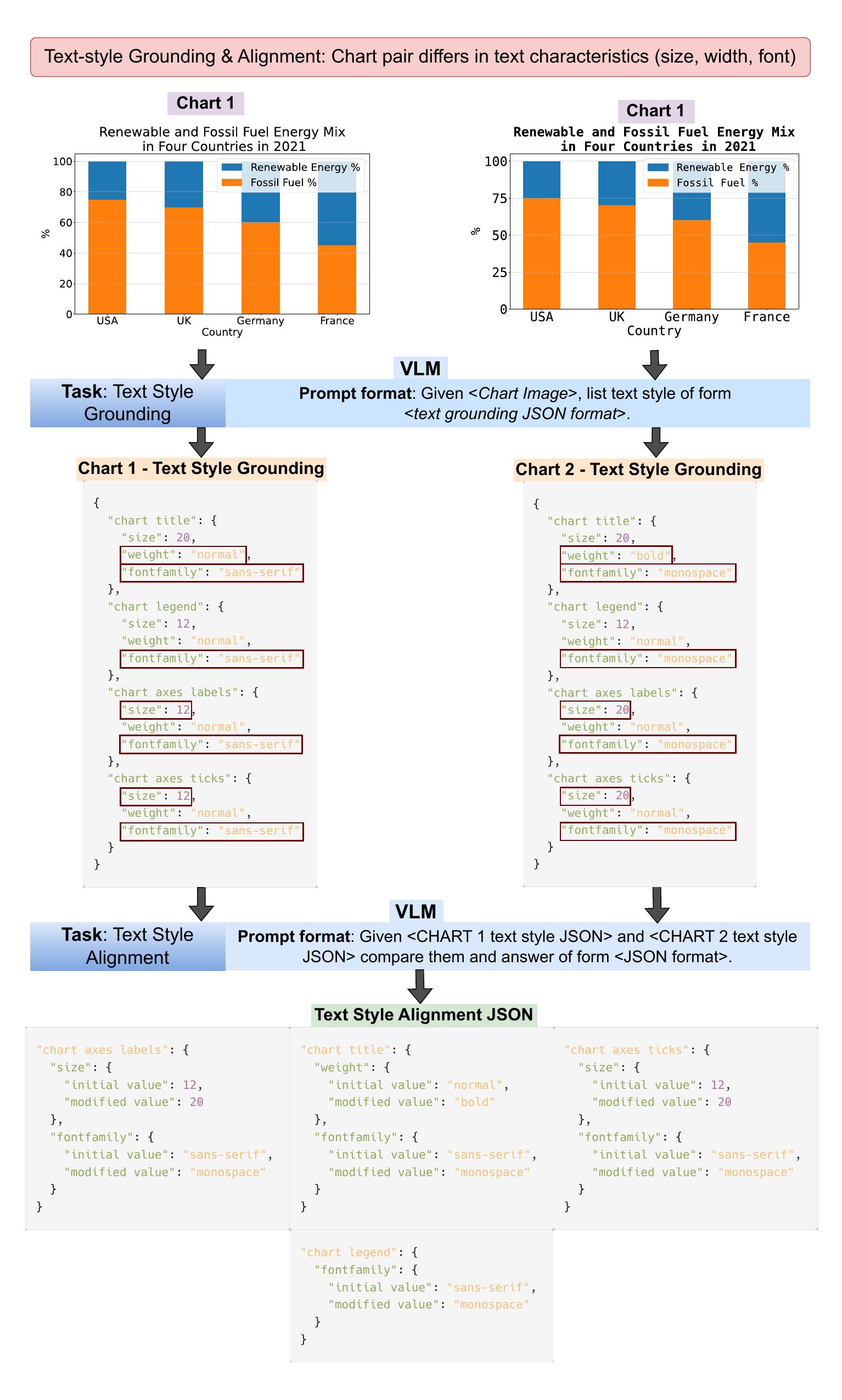}
   \vspace{-1em}
   \caption{\textbf{Two-Stage Evaluation Pipeline for \emph{Text Style Grounding \& Alignment} in \ours.} 
   The first stage focuses on grounding the text characteristics for the four chart regions: title, legend, axes labels, axes ticks. These characteristics are textual size, weight (lightness/boldness), and font family (e.g., Times New Roman). The second stage focuses on alignment, which aims to evaluate the grounded text characteristics and output a JSON file listing the characteristics for each region which differ between the chart pair.}
\label{fig:text_style_alignment_inference_pipeline}
\vspace{-1em}    
\end{figure*}

\begin{figure*}[htp]
\centering
\vspace{-1em}
   \includegraphics[width=0.85\linewidth]{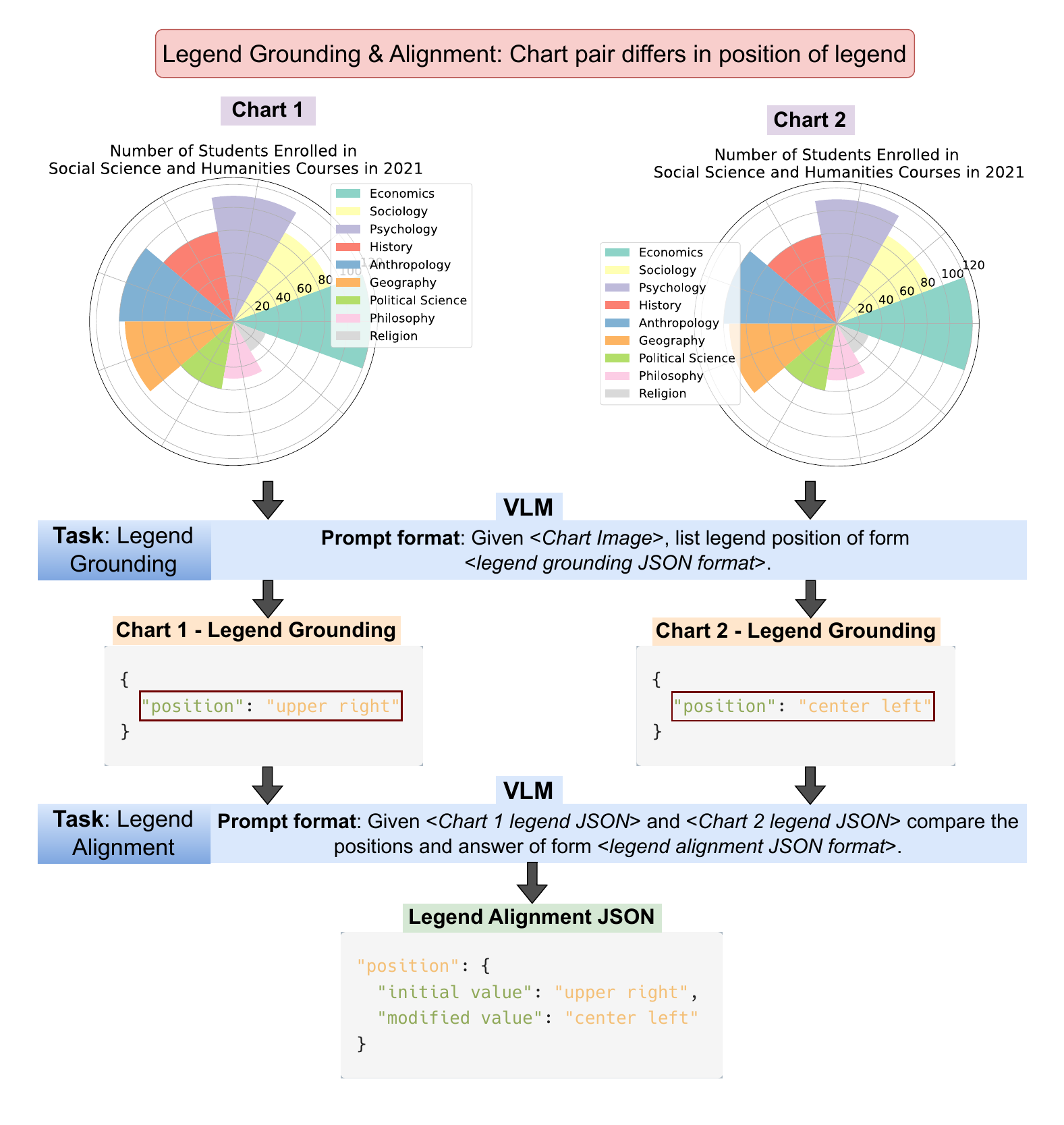}
   \vspace{-1em}
   \caption{\textbf{Two-Stage Evaluation Pipeline for \emph{Legend Grounding \& Alignment} in \ours.} 
   The first stage focuses on grounding the legend position in each chart, while the second stage focuses on alignment, which aims to determine the difference in the position and output the JSON file listing the difference. 
   }
\label{fig:legend_alignment_inference_pipeline}
\vspace{-1em}    
\end{figure*}

VLM outputs follow\textit{ JSON based formatting} due to precise nature of the key-value structure which is essential for element specific information serialization for finer-analysis, along with flexibility for variations in completion of grounding and fine grained analysis. The alignment JSON contains finer level attributes for which the charts differ, and the values for corresponding attribute in the two charts. E.g. for data alignment (as shown in Fig. \ref{fig:data_alignment_inference_pipeline}) the finer level attributes changed between the charts i.e. cells are identified by their row \& column header, along with its values in the chart pairs, i.e. value in chart 1 \& value in chart 2 respectively. Evaluation of attribute alignment tasks follow the same pipeline, as illustrated in Figure \ref{fig:color_alignment_inference_pipeline} for color alignment, Figure \ref{fig:text_style_alignment_inference_pipeline} for text-style alignment, Figure \ref{fig:legend_alignment_inference_pipeline} for legend alignment. 

% \includecomment{Why second stage? Why not programmatically evaluate grounding results? asked by Hongyu}
% \textit{Essentiality of second stage}: Second-stage forms essential part of evaluation pipeline. Analyzing dense-alignment ability requires performing end-to-end evaluation of VLMs. Grounding determines the chart information, and impacts the subsequent finer-level analysis. However correct grounding doesn't imply correct alignment. The VLM needs to make semantic correspondence between chart elements in the grounding result which is non-uniform and differs for each VLM. Moreover the hallucinating nature of VLMs make grounding output susceptible to ambiguities and vagueness, in which case the additional second-stage reasoning on the grounding result helps build a better overall understanding of VLM capabilities. Second-stage also allows utilization of additional contextual information (e.g. Chain-of-Thoughts) for the alignment task. Ultimately we analyze VLM's dense-alignment ability the way humans do looking at overall understanding, and at semantic shifts not captured by grounding.

\subsection{Evaluation metrics}
\label{app:evaluation_metric}

\includecomment{
1. JSON evaluation
2. JSON: key-value division
    - motivation
    - formalized for each task
3. key score (F1) + value score (precise)
4. task wise details
}

\subsubsection{Dense Alignment}
\label{app:evaluation_metric_align}
We evaluate dense alignment performance across four task categories: \textit{data alignment} (subtasks: 1-cell/2-cell/3-cell), \textit{color alignment}, \textit{text style alignment}, and \textit{legend alignment}. Performance on the first three tasks is evaluated by a key-value alignment score, which assess the capability to identify the different elements (keys) between two charts and their associated values. In contrast, 
% where we assess the model's ability to identify and localize specific chart elements that differ between image pairs. 
legend alignment score mainly focuses on identifying the different positions of legends in two charts (values only) because the key is unique and fixed. Table~\ref{tab:alignment_tasks} summarizes the keys and values of each type of elements as well as the notations of their dense alignment scores. 
% is evaluated separately based on spatial positioning accuracy.

\paragraph{Key-Value Alignment Score.}
For data, color, and text style alignment tasks, we define \textit{elements} as the atomic units that may differ across chart pairs. Each element is characterized by two components:
\begin{itemize}[leftmargin=*, itemsep=2pt]
    \item \textbf{Key}: A textual identifier that uniquely specifies the element within the chart.
    \item \textbf{Value}: The content or attribute value of the element in each chart of the pair.
\end{itemize}
The key serves to locate and identify different elements, while the values capture their corresponding data or content. 
% \paragraph{Alignment Score.}
We define the alignment score $s_{\text{align}}$ on a chart pair $(x, x')$ as:
\begin{equation}
s_{\text{align}}(x, x') = s_{\text{key}} + s_{\text{value}}
\end{equation}
where $s_{\text{key}} \in [0, 1]$ measures the key identification and $s_{\text{value}} \in [0, 1]$ measures the precision of predicted values. We rescale $s_{\text{align}}(x, x')$ to $[0,10]$ for better interpretability. 
% where each component contributes equally (5 points each). 
We will apply superscripts, e.g., $s^{(\text{data})}_{\text{align}}(x, x')$, to distinguish different task categorie, as shown in Table~\ref{tab:alignment_tasks}. 

\paragraph{Key Identification Score $s_{\text{key}}$} evaluates whether the model correctly identifies different elements between two charts. Let $K_{\text{gt}} = \{k_1, \ldots, k_n\}$ be the set of ground truth keys and $K_{\text{pred}} = \{\hat{k}_1, \ldots, \hat{k}_m\}$ be the set of predicted keys. We perform key matching between $K_{\text{gt}}$ and $K_{\text{pred}}$ using task-specific criteria: (1) for data and color alignment, we use Levenshtein distance with threshold $\tau = 0.5$ to account for the high lexical diversity of real-world named entities \citep{cohen2003comparison} and tabular headers \citep{zhang2019table2vec}; (2) for text style alignment, we require exact matches since the keys are predefined and region-characteristic. Let $K_{\text{valid}} = K_{\text{pred}} \cap_{\tau} K_{\text{gt}}$ denote the set of valid predicted keys, where $\cap_{\tau}$ represents the fuzzy intersection operator. We compute the following F1 score as $s_{\text{key}}$:
\begin{equation}
p_{\text{key}} = \frac{|K_{\text{valid}}|}{|K_{\text{pred}}|}, \quad
r_{\text{key}} = \frac{|K_{\text{valid}}|}{|K_{\text{gt}}|}, \quad
s_{\text{key}} = \frac{2 \cdot p_{\text{key}} \cdot r_{\text{key}}}{p_{\text{key}} + r_{\text{key}}}
\end{equation}

\paragraph{Precision of Predicted Values $s_{\text{value}}$.}
For each valid predicted element $k \in K_{\text{valid}}$, we measure the precision of its predicted values in both charts. Let $(v_k, v'_k)$ and $(\hat{v}_k, \hat{v}'_k)$ denote the ground truth and predicted values in charts $x$ and $x'$ respectively. The precision of predicted values is defined as
\begin{equation}
s_{\text{value}} = \frac{1}{2|K_{\text{valid}}|} \sum_{k \in K_{\text{valid}}} \left(\rho(v_k, \hat{v}_k) + \rho(v'_k, \hat{v}'_k)\right)
\end{equation}
where $\rho(\cdot, \cdot) \in [0, 1]$ denotes a task-specific value matching function: it performs exact matching for categorical attributes (e.g., text weight/font), $\rho(v, \hat{v}) = 1 - \|v - \hat{v}\|_2$ for color attributes (with $\|v - \hat{v}\|_2$ denoting the normalized RGB distance), and $\rho(v, \hat{v}) = 1 - \min(|v - \hat{v}| / |v|, 1)$ for numerical attributes (e.g., data values or text size).

% \[
% \rho_{\text{color}} = 1 - \tfrac{1}{2}\!\left(\tfrac{\|c_1 - \hat{c}_1\|_2}{d_{\max}} + \tfrac{\|c_2 - \hat{c}_2\|_2}{d_{\max}}\right),
% \]

% \paragraph{Task-Specific Instantiations.}
% \label{sec:task_instantiations}
% Table~\ref{tab:alignment_tasks} summarizes the element definitions and evaluation criteria for each dense alignment task.

\begin{table}[h]
\centering
\small
\resizebox{1.0\linewidth}{!}{\begin{tabular}{@{}llll@{}}
\toprule
\textbf{Task} & \textbf{Score} & \textbf{Key} & \textbf{Value} \\ 
\midrule
Data Alignment & $s_{\text{align}}^{\text(data)}(x, x')$ & Row and column labels & Numerical value \\
 & & (e.g., ``John, Salary'') & (float/int) \\
\midrule
Color Alignment & $s_{\text{align}}^{\text(color)}(x, x')$ & Series/category label & Hex color code \\
 & & (e.g., ``Product A'') & (e.g., ``\#FF5733'') \\
\midrule
Text Style Alignment & $s_{\text{align}}^{\text(text-style)}(x, x')$ & Region-characteristic pair & Style attribute value \\
Alignment &  & (e.g., ``title-size``) & (size: int, weight/family: categorical) \\
\midrule
Legend Alignment & $s_{\text{align}}^{\text(legend)}(x, x')$ & Position (implicit) & 3X3 grid (center, upper, ...) \\
 & & & \\
\bottomrule
\end{tabular}}
\caption{Chart elements' keys, values, and scores in the four categories of dense alignment tasks. For data, color, and text style alignment, fuzzy matching (Levenshtein distance $\tau=0.5$) or exact matching is used to evaluate the key identification, while the precision of associated values are evaluated using $\rho(\cdot, \cdot)$. Legend alignment score is defined by spatial distance between the values of legend positions.}
\label{tab:alignment_tasks}
\end{table}

\paragraph{Legend Alignment Score.}
Unlike the above three alignment tasks, legend alignment only focuses on one unique key, i.e., the legend position, so the legend alignment score is defined as the spatial proximity between the ground truth and model-detected positions. We discretize the chart into a $3 \times 3$ grid and measure the Manhattan distance between predicted and ground truth legend positions. The legend alignment score is defined by
\begin{equation}
s^{\text{(legend)}}_{\text{align}}(x, x') = 1 - \frac{1}{10} \cdot ( d_{\text{Manhattan}}({pos}, \hat{pos}) + d_{\text{Manhattan}}(pos', \hat{pos'}))
\end{equation}
where $\hat{pos}$ and $pos$ are the predicted and ground truth positions, and $d_{\text{Manhattan}}(\cdot, \cdot) \in [0, 5]$ is the Manhattan distance. We normalize $s^{\text{(legend)}}_{\text{align}}(x, x')$ to $[0,10]$ for better interpretability.

For each chart type, we report the averaged alignment scores over all the chart pairs belonging to that chart type.

\subsubsection{Robustness}
\label{app:evaluation_metric_robust}
\paragraph{} We evaluate the robustness of data alignment performance to the variations of visual attributes. For a chart pair $(x, x')$ differing in a 1-3 data cells, we define robustness $r(x, x')$ as the reciprocal of the standard deviation $\sigma(\cdot)$ of alignment scores across $d$ visual variations:
\begin{equation}
r(x, x') = \frac{1}{1 + \sigma\left(\left\{s^{(\text{data})}_{\text{align}}(x^{(j)}, x'^{(j)})\right\}_{i=1}^{d}\right)}
\end{equation}
where $(x^{(j)}, x'^{(j)}), \ldots, (x^{(d)}, x'^{(d)})$ are the $d$ visually-varied versions of the same chart pair $(x,x')$, and $s^{(\text{data})}_{\text{align}}$ denotes the data alignment score. Higher $r(x, x')$ indicates more consistent data alignment performance across different visual variations. We compute robustness separately for each attribute $a \in \{\text{color}, \text{legend}, \text{text style}\}$. 
For each chart type, we report the robustness score averaged over all the chart pairs belonging to that chart type. 

\subsection{Additional Experimental Details}

\subsubsection{VLM Selection}
\label{app:vlm_selection}

We evaluate a diverse suite of \textit{open-source VLMs} from following families: Phi-3.5 vision-instruct \cite{abdin2024phi}, InternVL-2.5 (8B) \cite{chen2024expanding}, LLaVA-1.6 Mistral (7B) \cite{liu2023improved}, QWEN-2.5 VL (8B) \cite{bai2025qwen2}. These models constitute among most widely used VLMs, and have a long timeline of continuous evolution with each released version. The set encompasses the top-performed VLMs in various chart benchmarks (CharXiv \cite{wang2024charxiv}, ChartQAPro \cite{masry2025chartqapro}, SCI-CQA \cite{li2023scigraphqa}, MultiChartQA \cite{zhu2024multichartqa}, discussed in \ref{related_work:chart_understanding_benchmarks}).
% ChartBench \cite{xu2023chartbench}
% Recent CharQAPro \cite{masry2025chartqapro} benchmark also focus on evaluating exactly the same four open-source models (Fig 1: ChartQA vs ChartQAPro comparison), mainly because of their outstanding capability of chart understanding.

% \dang{It would be more impactful if you can include more models since this is a benchmark paper. For example, GPT-4.1, o4-mini, GPT-5, Claude 4 Sonnet, etc.}

Our choice of \textit{proprietary VLM} is based on CharXiv \cite{wang2024charxiv} leaderboard as its tasks/questions require dense-level grounding. For example, CharXiv tasks need to identify axes ticks by positions and their value enumerartion, grid-lines count and intersections, integral (area comparison of regions) and slope (rate of increase/decrease) in line charts.
And GPT-4o \cite{hurst2024gpt} is the best performing proprietary in the CharXiv paper. 

Among \textit{chart-specialized VLMs}, we evaluate TinyChart \cite{zhang2024tinychart} \& ChartGemma \cite{masry2024chartgemma} models. However, due to their task-specific training (discussed in \ref{related_work:vlm_for_charts}), these models show collapse of instruction following capabilities and fail to output required JSON format needed for evaluation. Below are a few examples of the outputs.

JSON output: Data alignment (1 cell) by ChartGemma and TinyChart models using 1-stage stitched-charts (i.e chart pair stacked as single image) evaluation.

\begin{verbatim}
REQUIRED FORMAT (specified in prompt instructions):-
{"row name": <row name of the cell>, "column name": <column name of the cell>,
 "value in chart 1": <value in first chart of the pair>, "value in chart 2": 
 <value in second chart of the pair>}

EXAMPLE:-
{"row name": "Production A (million units)", "column name": "2021",
 "value in chart 1": 35, "value in chart 2": 30}

CHARTGEMMA OUTPUT (abnormal valued JSON which is inconsistent with required format):-
{"row name": "sample row", "column name": "sample column",
 "value in chart 1": Infinity, "value in chart 2": Infinity}

TINYCHART OUTPUT (abnormal list instead of JSON):-
["Production A (million units)", "Production B (million units)",
 "Production C (million units)" ..... "Production Z (million units)"]
\end{verbatim}

\subsubsection{Ablations}
\label{app:ablation_study}

\begin{table*}[ht]
\small
\centering
\resizebox{\textwidth}{!}{%
    \begin{tabular}{llccccccccc}
    \toprule
    \textbf{Type} & \textbf{Approach} & \textbf{Bar} & \textbf{Bar \#} & \textbf{3D Bar} & \textbf{Line} & \textbf{Line \#} & \textbf{Radar} & \textbf{Rose} & \textbf{Box} & \textbf{Multi-Axes} \\
    \midrule
    \multirow{2}{*}{1-stage} 
        & Multi-chart       & 2.6 & 4.5 & 1.9 & 2.9 & 3.0 & 1.1 & 0.1 & 0.9 & 0.8 \\
        & Stitched-chart    & 2.1 & 2.2 & 0.8 & 1.9 & 0.9 & 0.5 & 0.1 & 0.1 & 0.4 \\
    \cmidrule{1-11}
    2-stage & Ours           & 4.7 & 7.0 & 1.7 & 5.4 & 5.9 & 1.0 & 0.1 & 0.4 & 0.7 \\
    \bottomrule
    \end{tabular}
}
% \resizebox{\textwidth}{!}{%
%     \begin{tabular}{llccccccccc}
%     \toprule
%     \textbf{Type} & \textbf{Approach} & \textbf{Bar} & \textbf{Bar \#} & \textbf{3D Bar} & \textbf{Line} & \textbf{Line \#} & \textbf{Radar} & \textbf{Rose} & \textbf{Box} & \textbf{Multi-Axes} \\
%     \midrule
%     \multirow{2}{*}{1-stage} 
%         & Multi-chart       & 4.8 & 7.4 & 4.7 & 3.3 & 4.7 & 4.9 & 3.1 & 3.2 & 3.3 \\
%         & Stitched-chart    & 5.0 & 4.8 & 3.0 & 4.5 & 3.5 & 3.0 & 2.7 & 2.8 & 3.2 \\
%     \cmidrule{1-11}
%     2-stage & Ours           & 6.5 & 8.3 & 4.1 & 6.1 & 6.3 & 3.8 & 3.4 & 2.9 & 3.5 \\
%     \bottomrule
%     \end{tabular}
% }
\caption{\textbf{Ablation study of 1-stage vs. 2-stage evaluations} on data alignment (one cell change) task. Mean scores across nine chart types show that our 2-stage evaluation reflects VLMs' greatest potential on chart alignment.}
\label{tab:Ablation table}
\end{table*}

% \dang{Ideally, a large table/figure comparing different VLMs on our proposed benchmark should appear first, followed by a discussion of trends, for example, perception biases or weakness patterns in current VLMs. Ablations and other analyses should then follow.}
We performed ablation experiments to vigorously compare differing approaches to our 2-stage approach.

The ablation experiments aimed to thoroughly compare single-stage based alignment approaches for performing multi-image reasoning vis-a-vis our two-stage approach. The ablation techniques:-

(1) \textit{stitched-charts} inference: The chart-pair images are vertically concatenated resulting in a single image of stitched chart-pairs which undergo single-stage inference. 

(2)\textit{ multi-image} inference: The VLM inputs multiple images, and contextualizes output based on the input images with aim of better understanding across of finer-level alignment in multi-image reasoning.

The ablation experiments evaluated the Phi-3.5 model’s performance on the data alignment task.
As shown in Table~\ref{tab:Ablation table}, the single-stage approach underperformed compared to our proposed two-stage method, reaffirming the effectiveness of intermediate grounding for reasoning in the alignment task, helping to focus more precisely on localized relationships between visual and textual elements. In contrast, the single-stage approach struggles to capture these fine-grained correspondences due to information loss during joint encoding and limited cross-attention resolution. Despite continued progress in multi-modal training, current VLMs still face challenges in detailed reasoning, and our results highlight how decomposing complex tasks like our multi-chart dense alignment into modular stages can substantially mitigate these limitations.

\includecomment{general > finer, grounding charts restricted by grounding}
% The chart-types results show major gaps in model's finer level understanding especially for general charts which are assumed to 
% Intuitively we may expect two-stage to more favorably impact the difficult aspects on which single-stage approaches lack, however the results show opposing trend 

\subsection{Additional Finding \& Insights}

\begin{figure*}[htbp]
    \centering
    \vspace{-1em}
    
    \begin{subfigure}[t]{0.48\textwidth}
        \includegraphics[width=\linewidth]{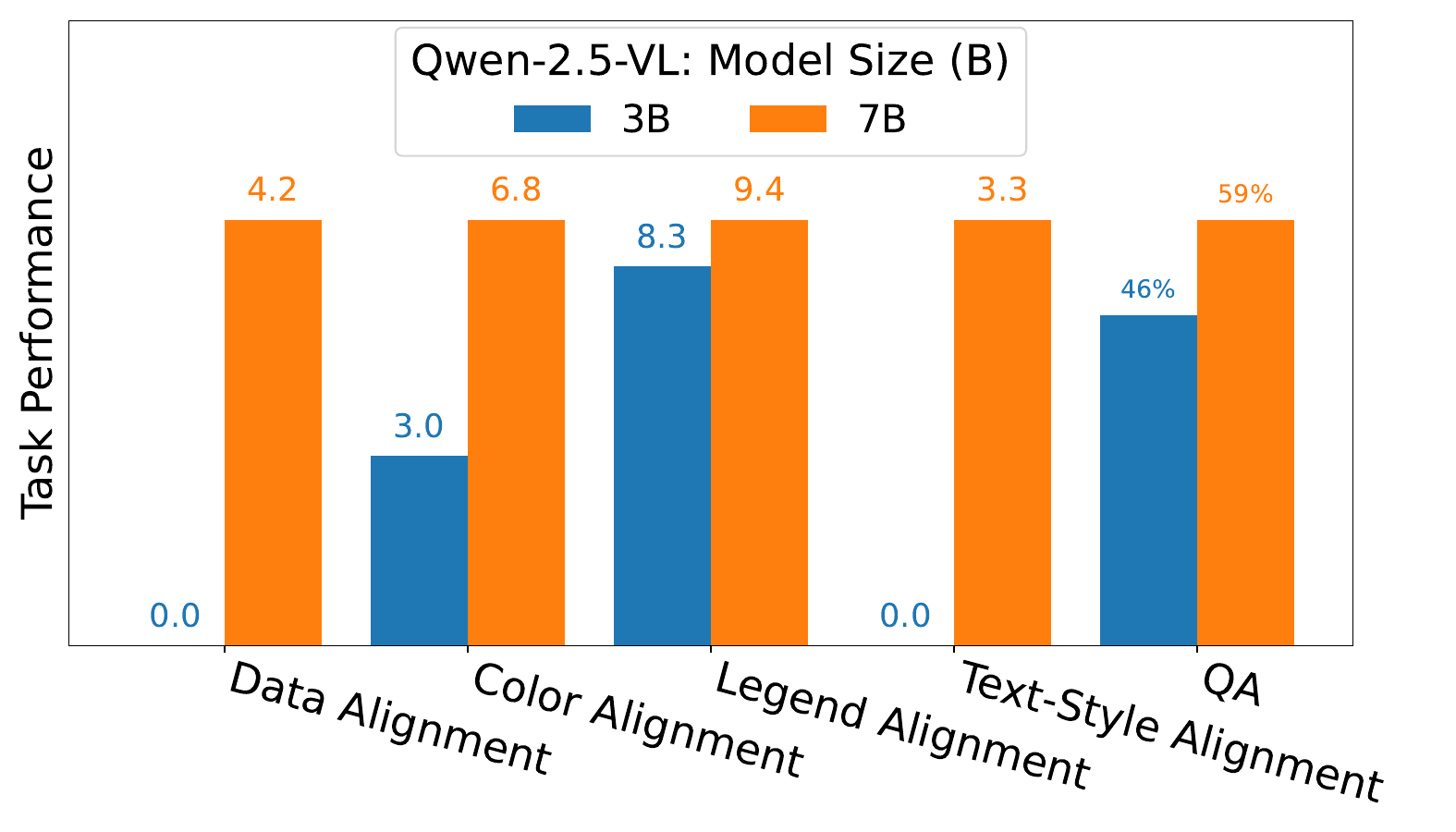}
       \end{subfigure}
    \hfill
    \begin{subfigure}[t]{0.48\textwidth}
        \includegraphics[width=\linewidth]{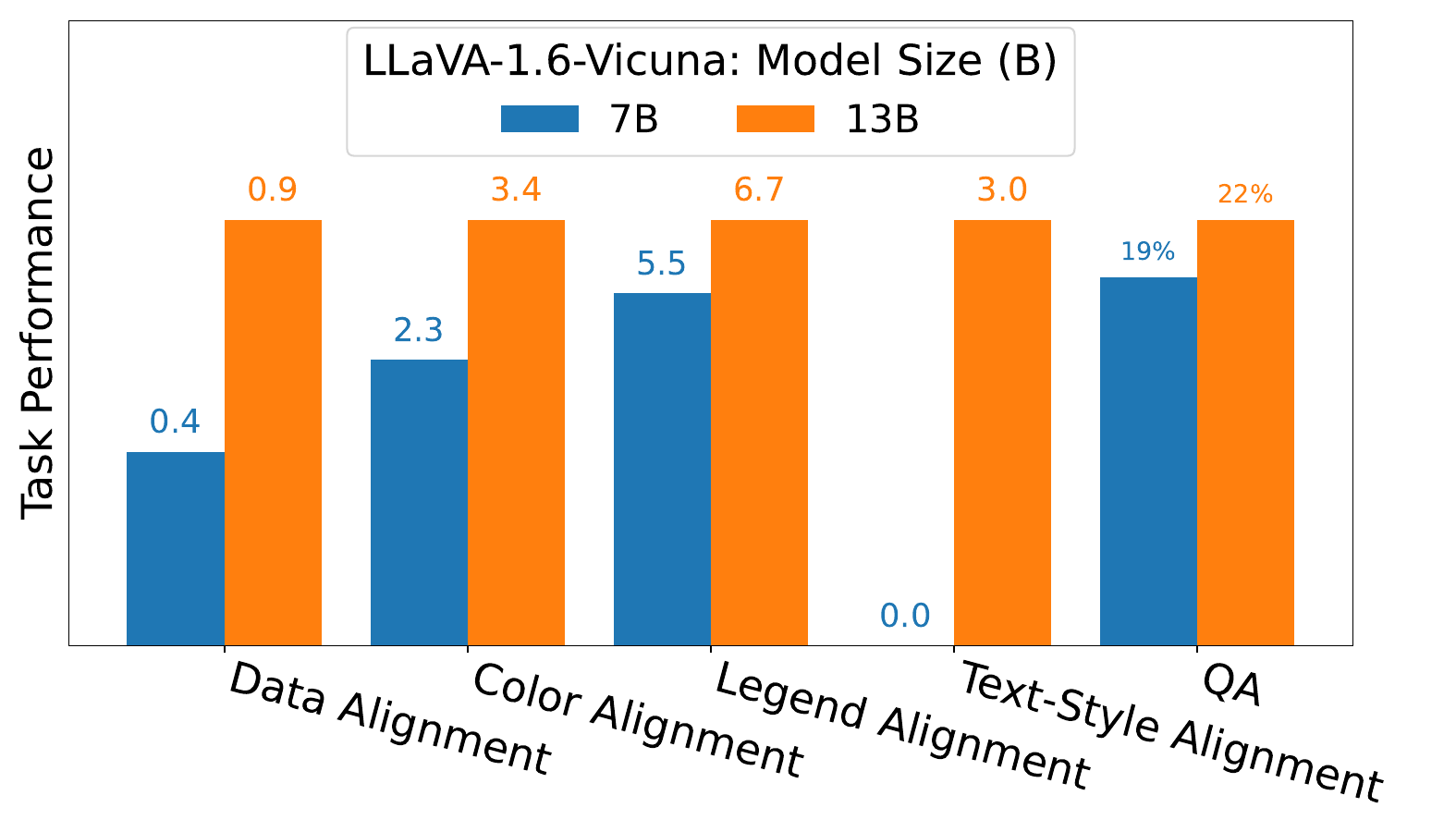}
    \end{subfigure}
    
    \caption{Task performances for different sizes of Qwen-2.5-VL and LlaVa-Vicuna-1.6.}
    \label{fig:scaling_law_qwen_and_llava}
\end{figure*}

\begin{figure*}[htbp]
    \centering
    \vspace{-1em}
    
    \begin{subfigure}[t]{0.48\textwidth}
        \includegraphics[width=\linewidth]{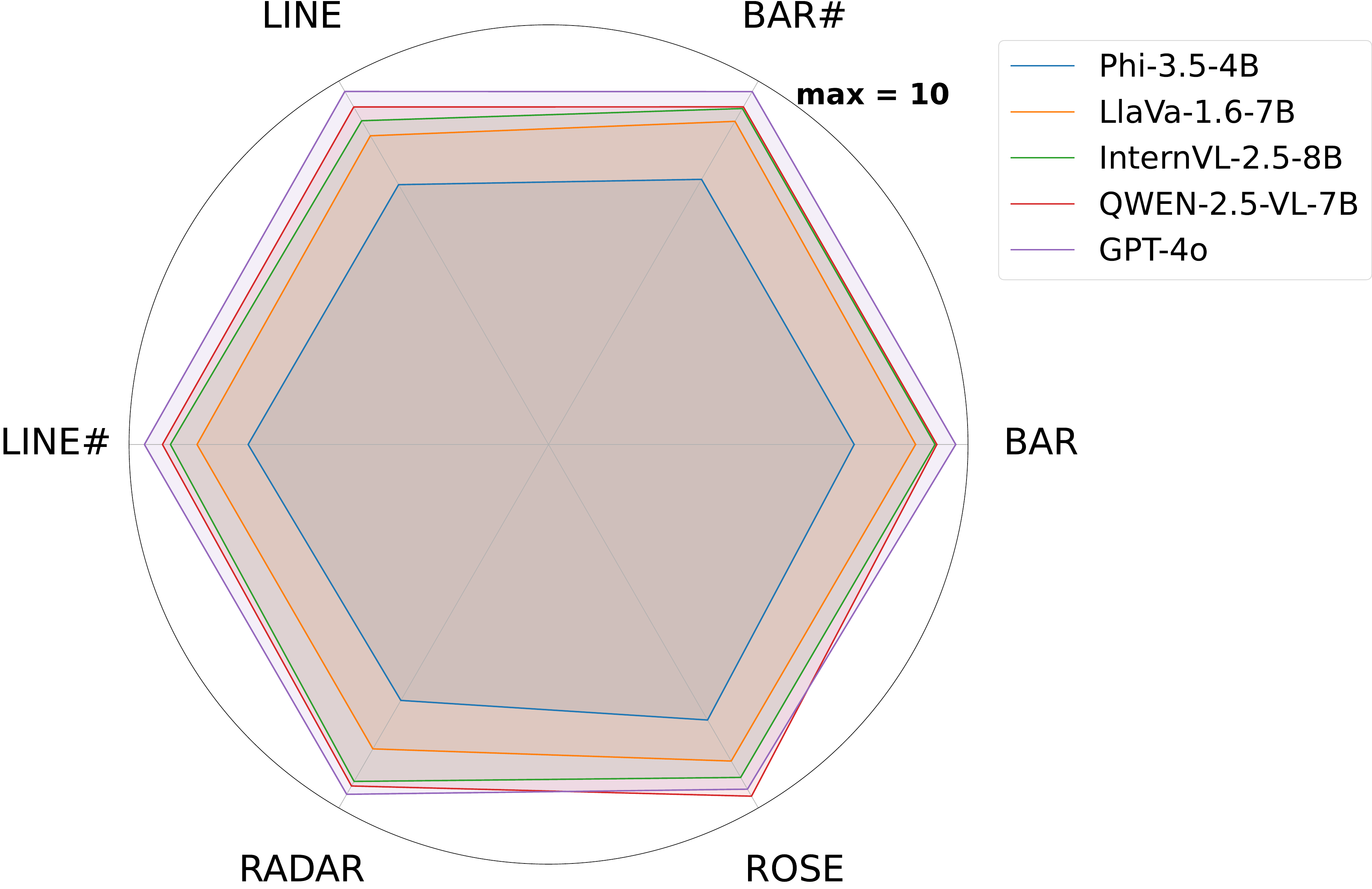}
        \caption{Legend Alignment}
        \label{fig:legend_alignment_radar_charts}
    \end{subfigure}
    \hfill
    \begin{subfigure}[t]{0.48\textwidth}
        \includegraphics[width=\linewidth]{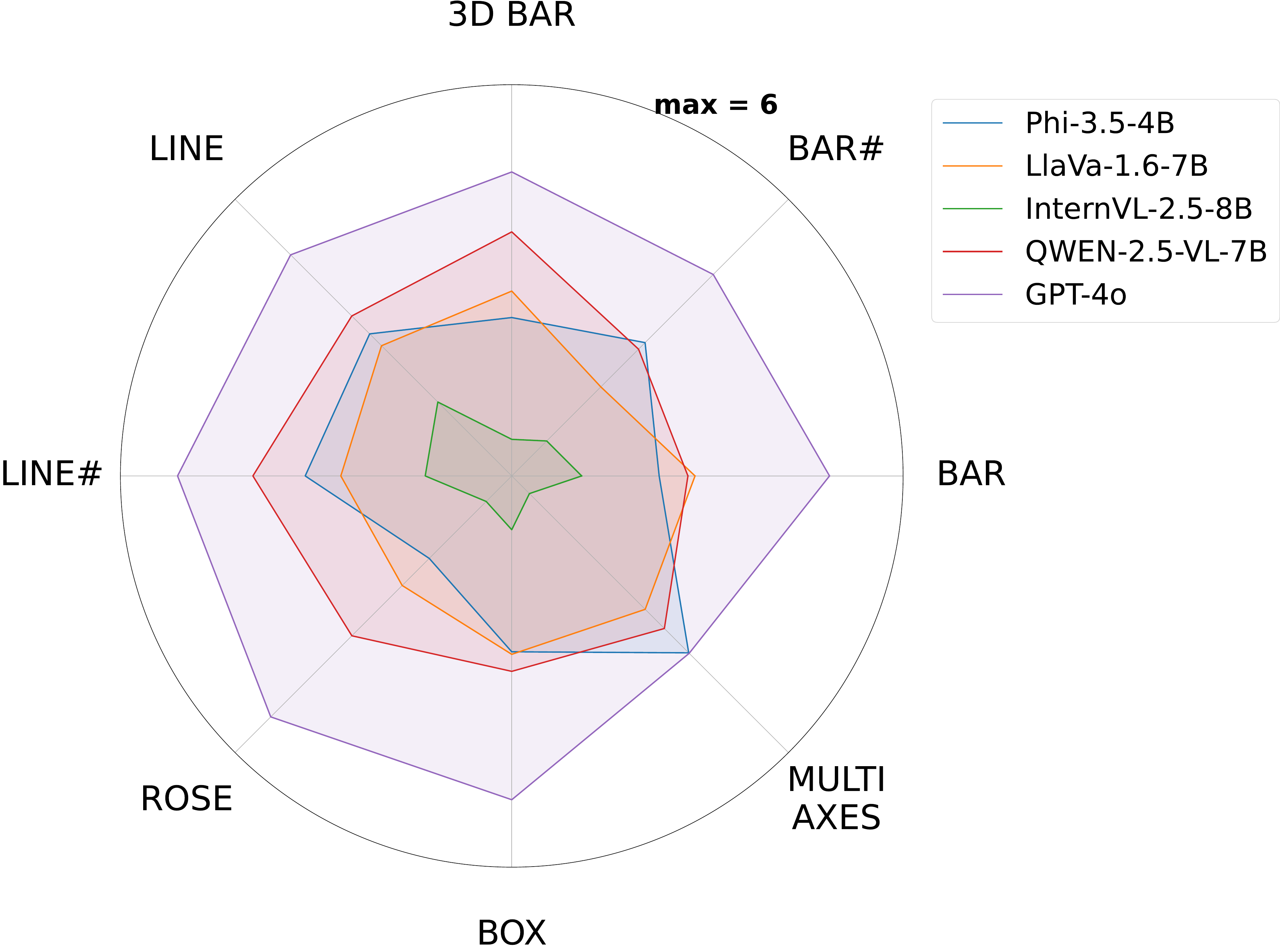}
        \caption{Text-Style Alignment}
        \label{fig:text_style_alignment_radar_charts}
    \end{subfigure}
    
    \vspace{-0.5em}
    \caption{(a) \textbf{Legend alignment} of legend positions. Phi-3.5 performs the worst while GPT-4o is best. Related discussion in Finding 1\&2. (b) \textbf{Text-style alignment} (size, weight, font). Worst: InternVl-2.5-8B, Best: GPT-4o. Discussion in Finding 1\&4.}
    \label{fig:attr_alignment_comparison}
    \vspace{-1em}
\end{figure*}

\begin{figure}[htp]
\centering
% \vspace{-1em}
   \includegraphics[width=1.0\linewidth]{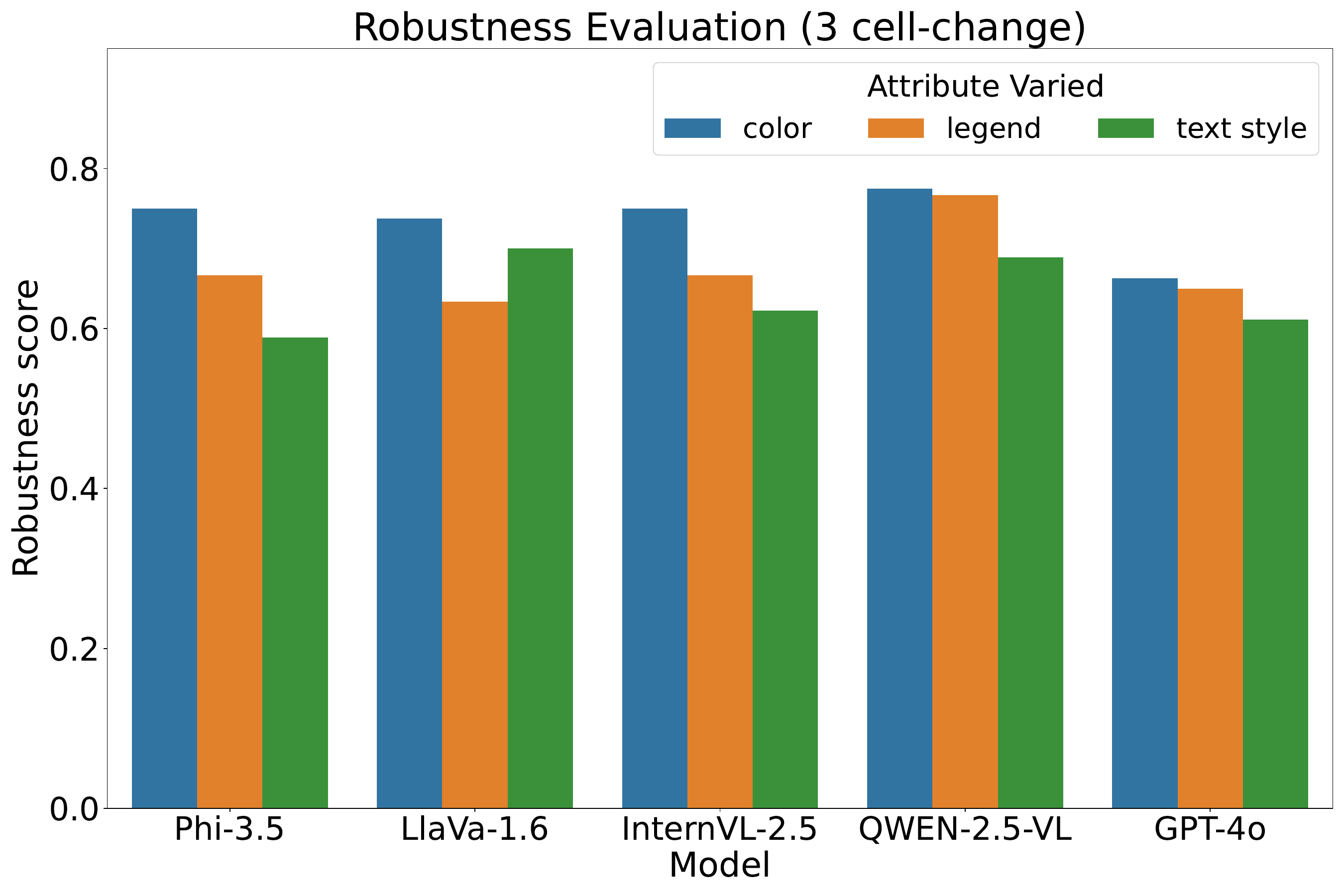}
   \vspace{-1em}
   \caption{\textbf{VLMs' Robustness of data alignment (3-cell change) to variations in color, legend, and text-style.} VLMs show better robustness to color changes than text-style changes. QWEN-2.5-VL outperforms the other four VLMs on robustness. More discussion can be found below Finding 6.
   }
   \vspace{-1em}
    \label{fig:robustness_results_bar_plots}
\end{figure}

% \paragraph{P5. Model's robustness in data alignment}
  % \end{subfigure}
\begin{findingbox}[title={Finding 7}]
% \label{par:}
VLMs' data grounding and alignment are more robust to color variations than changes in legend positions and text styles. 
\end{findingbox}
Fig. \ref{fig:robustness_results_bar_plots} shows that robustness is the worst under text-style variations and the best under color variations.  
% We observe least robustness for text-style variation while most for color variation. (Fig. \ref{fig:robustness_results_bar_plots}). Unlike color in user perception focused on engagement and storytelling, 
In the visualizations of data, colors are used to discretize, categorize, and measure chart constituents. As long as their colors are distinguishable, color variations will not affect the data grounding. 
In contrast, the text styles and legends provide critical information about the data via ticks, labels, and legend items. Moreover, changing legend position may lead to position changes and occlusion of other chart elements. Hence, their variations have a greater impact on the data grounding/alignment performance.  
% ence structured nature and clear constituent partitioning enable data-grounding for VLMs' under reasonable shade difference. The chart text (including ticks, legend, labels) strongly drives VLM's understanding of chart data. We hypothesize text-style influence on value (represented by text) alignment vis-a-vis corresponding chart characteristic unlike default used text-styles hence making the data alignment and grounding imprecise.
% The chart constituent positions being important for data grounding, legend position change tends to influence remaining constituent's positions and their relative ordering hence leading to less robustness compared to color variations. 

\begin{figure}
  \centering
  \begin{subfigure}[b]{0.4\textwidth}  
    \centering
    \includegraphics[width=\textwidth]{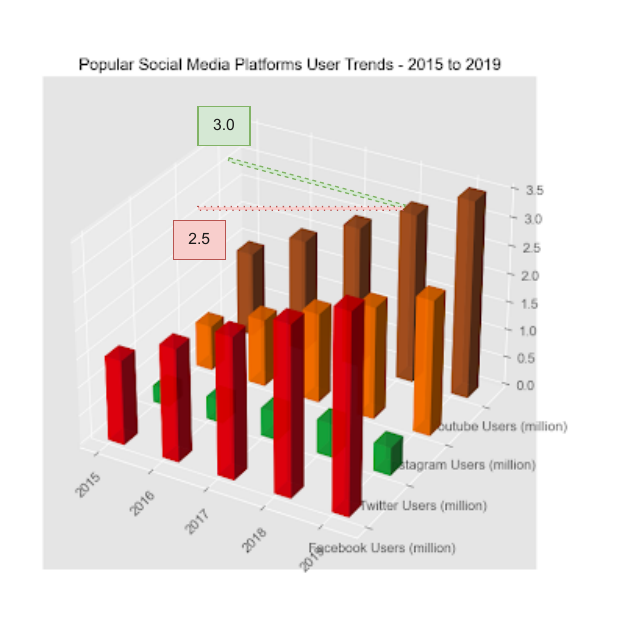}
    \caption{Depth estimation in 3D bar charts}
\label{fig:3d_chart_depth_estimation_example}
\end{subfigure}
  \hfill
  \begin{subfigure}[b]{0.4\textwidth}
    \centering
    \includegraphics[width=\textwidth]{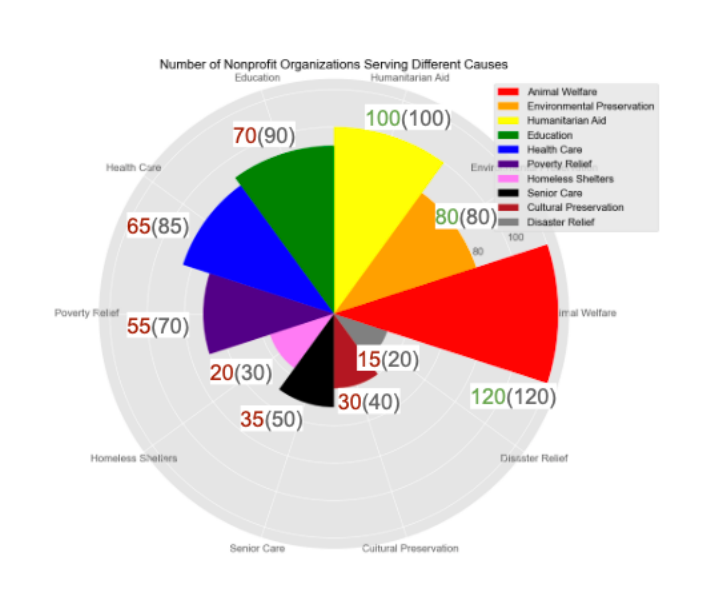}
    \caption{Text vs. non-text cues for value scaling in rose charts.\looseness-1}
    \label{fig:rose_chart_text_vs_grid_example}
  \end{subfigure}
  \vspace{-1em}
  \caption{\textbf{VLMs' spatial understanding is poor on complex charts.} More discussion is provided below Finding 7.}
  \label{fig:spatial_numerical_mapping}
\end{figure}

% \paragraph{P6. Spatial Understanding}
\begin{findingbox}[title={Finding 8}]
% \label{par:}
VLMs' spatial understanding capability affects several important chart understanding skills. 
\end{findingbox}
Chart understanding usually requires an accurate mapping between spatial relationships and the corresponding numerical values to be visualized. 
% spatial-numerical mapping ability which drives chart data understanding, shows deficit for required nuanced understanding. 
\begin{itemize}
    \item \textit{Depth understanding}: Despite the high-level similarity between 3D bar charts and (2D) bar charts, as shown in Fig~\ref{fig:data_and_color_alignment_radar_plot}, the data alignment performance is much poorer on 3D bar charts due to the lack of depth understanding, which affects the measurement of scales and values along axes in the 3D space.   
    % The 3D bar charts despite being similar to bar chart in high level structure and attributes has drastic data-alignment difference. (Fig \ref{fig:data_alignment_1_cell_model_comparison_radar_chart}). This can be attributed to VLM's inefficient depth understanding. Along with the axes and precise value estimation for understanding scaling \& estimating values. Hence resulting in frail data-alignment in 3D bar compared bar chart.
    \item \textit{Text vs non-text cues}: Rose charts are extended from bar charts by allowing more polar coordinates with scale differences in radial forms. However, Fig.~\ref{fig:rose_chart_text_vs_grid_example} reveals a great difference between the two on data alignment performance. 
    This is due to fewer text cues (e.g., axes ticks) in rose charts, where non-text cues such as grid lines cannot be fully leveraged.  
    % The rose chart result from bar-chart adapted to polar-coordinate representation to show-case scale differences in radial form. 
    % The underlying gaps in scale understanding from absence of text cues (i.e. all axes ticks) while inability of non-text cues (grid-lines) to replace text-based understanding (Fig. \ref{fig:rose_chart_text_vs_grid_example}) leads to weaker data alignment in rose chart compared to bar chart.
    \item \textit{Better performance on numbered charts}: numbered bar and line charts explicitly place the data values in the charts, hence facilitating VLMs to extract the data easily without precise measurements of the visual elements. Hence, as shown in Fig.~\ref{fig:data_and_color_alignment_radar_plot}, numbered bar/line charts usually enjoy better performance. 
    % The bar-numbered chart and line-numbered chart augment data information by labeling precise encoding-wise value in bar chart and line chart respectively. Hence enabling models to directly extract associated data value hiving axes scale understanding ability. Thus resulting in numbered charts performing un-numbered compatriots for each of the models. (Fig. \ref{fig:data_alignment_1_cell_model_comparison_radar_chart})   
\end{itemize}

% \begin{figure}
%   \centering
%   \begin{subfigure}[b]{0.4\textwidth}  
%     \centering
%     \includegraphics[width=\textwidth]
% \end{subfigure}
%   \hfill
%   \begin{subfigure}[b]{0.4\textwidth}
%     \centering
%     \includegraphics[width=\textwidth]{imgs/rose_chart_example.drawio.pdf}
%     \caption{Text vs. non-text cues for value scaling in rose charts.\looseness-1}
%     \label{fig:rose_chart_text_vs_grid_example}
%   \end{subfigure}
%   \vspace{-1em}
%   \caption{\textbf{VLMs' spatial understanding is poor on complex charts.} More discussion is provided below Finding 7.}
%   \label{fig:spatial_numerical_mapping}
% \end{figure}

\end{document}